**IEEE** *Access*

Multidisciplinary | Rapid Review | Open Access Journal

## SURVEY

# Comprehensive Overview of Reward Engineering and Shaping in Advancing Reinforcement Learning Applications

**SINAN IBRAHIM** [1], **MOSTAFA MOSTAFA** [1], **ALI JNADI** [2,3], **HADI SALLOUM** [3], **AND PAVEL OSINENKO** [1]

[1] Skolkovo Institute of Science and Technology, 121205 Moscow, Russia
[2] Institute of Robotics and Computer Vision, Innopolis University, 420500 Innopolis, Russia
[3] Research Center for Artificial Intelligence, Innopolis University, 420500 Innopolis, Russia

Corresponding author: Sinan Ibrahim (Sinan.Ibrahim@Skoltech.ru)

The work of Ali Jnadi and Hadi Salloum was supported by the Analytical Center for the Government of Russian Federation under Agreement 70-2021-00143 01.11.2021 and Agreement IGK 000000D730324P540002.

**ABSTRACT** Reinforcement Learning (RL) seeks to develop systems capable of autonomous decision-making by learning through interaction with their environment. Central to this process are reward engineering and reward shaping, which are essential for enhancing the efficiency and effectiveness of RL algorithms. These techniques guide agents toward desired behaviors, improve learning stability, and accelerate convergence by addressing challenges such as sparse and delayed rewards. However, the complexity of real-world environments and the computational demands of RL algorithms remain significant obstacles to broader adoption. Recent advancements in deep learning have enabled RL to handle high-dimensional state and action spaces, facilitating applications in robotics, autonomous driving, and complex decision-making tasks. In response to these developments, this paper provides one of the first comprehensive reviews of reward design in RL, with a focus on the methodologies and techniques underpinning reward engineering and shaping. By introducing a detailed taxonomy, critically analyzing current approaches, and highlighting their limitations, this work fills an important gap in the literature, offering insights into how reward structures can be optimized to meet the growing demands of modern AI systems.

## I. INTRODUCTION

Reward design in Reinforcement Learning (RL) is a critical component that significantly influences the performance and learning efficiency of RL agents [1], [2], [3]. RL, a prominent subset of Machine Learning, trains intelligent agents to make sequential decisions by learning from interactions with their environment [4]. These agents aim to maximize cumulative rewards over time, making the design of the reward function pivotal to their success. Reward design is a nuanced and intricate process that involves defining the reward function in a way that aligns with the desired behavior and goals

The associate editor coordinating the review of this manuscript and approving it for publication was Huiyan Zhang.

of the RL agent. As highlighted in [5], the importance of reward design cannot be overstated, as it directly impacts the agent's ability to learn and adapt to complex environments. The art of reward design can be categorized into two primary areas. The first is Reward Engineering [6] which involves the creation of the reward function itself. This involves the initial creation of the reward function. The reward function, $R(s, a, s')$, maps states $s$, actions $a$, and successor state $s'$ to a numerical reward value. A well-designed reward function should provide informative feedback to the agent, guiding it toward desired actions and behaviors. The reward function must strike a balance between being informative enough to facilitate learning and sparse enough to prevent trivial solutions. For example, in a grid-world navigation task, the reward function might be defined as:

$$R(s, a, s') = \begin{cases} +10 & \text{if } s' \text{ is the goal state} \\ -1 & \text{if } s' \text{ is a non-goal state.} \end{cases}$$

The reward function [3], [7], [8] serves as a signal to the RL agent, guiding it towards desirable actions and behaviors. A well-designed reward function should be informative, providing the agent with clear feedback on the quality of its actions. It should also be sparse enough to prevent trivial solutions but dense enough to facilitate learning. For example, in a game-playing scenario, reward engineering might involve assigning positive rewards for winning moves and negative rewards for losing moves, with additional considerations for intermediate actions that contribute to the overall strategy.

The second category is Reward Shaping [9], [10]. Once the reward function is established, reward shaping comes into play to fine-tune and enhance the reward signals. Reward shaping involves modifying the reward function to improve the learning process without altering the optimal policy. One common approach is potential-based reward shaping [11], where a potential function $\Phi(s)$ is introduced to modify the rewards:

$$R'(s, a, s') = R(s, a, s') + \gamma \Phi(s') - \Phi(s). \quad (1)$$

Here, $\gamma$ is the discount factor. The potential function $\Phi(s)$ should be carefully designed to ensure it does not change the optimal policy but accelerates learning by providing intermediate rewards. For instance, in a robotic arm manipulation task, $\Phi(s)$ could represent the negative distance to the target object, encouraging the agent to move closer to the target. Techniques such as potential-based shaping functions can be used to provide additional guidance to the agent, accelerating the learning process by giving intermediate rewards for progress toward the ultimate goal. For instance, in a navigation task, shaping rewards could include providing positive feedback for reaching sub-goals or making progress towards the destination, even if the final goal is not yet achieved.

Effective reward design requires a deep understanding of the problem domain, the agent's learning dynamics, and potential pitfalls such as reward hacking, where the agent finds unintended shortcuts to maximize rewards without achieving the desired behavior. To mitigate such problems, iterative testing and validation of the reward function are essential, ensuring that the designed rewards lead to the intended outcomes. Reward design in RL is a fundamental aspect that encompasses both the creation and refinement of reward functions. Through careful reward engineering and shaping, one can guide RL agents to learn complex behaviors efficiently and effectively, ultimately leading to the development of intelligent systems capable of performing sophisticated tasks.

For deep understanding, consider an RL problem formulated as a Markov Decision Process (MDP), defined by the tuple $(S, A, P, R, \gamma)$, where:

- $S$ is the set of states,
- $A$ is the set of actions,
- $P(s'|s, a)$ is the transition probability function,
- $R(s, a)$ is the reward function,

The goal of the RL agent is to learn a policy $\pi(a|s)$ that maximizes the expected cumulative reward, given by the return $G_t$:

$$G_t = \sum_{k=0}^{\infty} \gamma^k R(s_{t+k}, a_{t+k}). \quad (2)$$

The value function $V^\pi(s)$ and the action-value function $Q^\pi(s, a)$ are defined as:

$$V^\pi(s) = \mathbb{E}_\pi \left[ \sum_{k=0}^{\infty} \gamma^k R(s_{t+k}, a_{t+k}) \Big| s_t = s \right], \quad (3)$$

$$Q^\pi(s, a) = \mathbb{E}_\pi \left[ \sum_{k=0}^{\infty} \gamma^k R(s_{t+k}, a_{t+k}) \Big| s_t = s, a_t = a \right]. \quad (4)$$

In reward shaping, the modified reward function $R'(s, a, s')$ ensures the policy remains optimal while improving the learning speed. The shaping potential $\Phi(s)$ is chosen such that the difference in potential provides informative guidance:

$$\Phi(s) = \text{heuristic}(s). \quad (5)$$

For instance, in a maze-solving task, $\Phi(s)$ could be the negative Manhattan distance to the goal, providing incremental rewards as the agent approaches the goal.

Reward design in RL is a fundamental aspect that encompasses both the creation and refinement of reward functions and it has been used in various areas [12], [13], [14], [15]. Through careful reward engineering and shaping, one can guide RL agents to learn complex behaviors efficiently and effectively, ultimately leading to the development of intelligent systems capable of performing sophisticated tasks. Efficiently, the complexity of modern systems demands a more nuanced approach to reward engineering. The work cited as [6] argues for a shift in focus toward developing alternative frameworks that can effectively guide these agents toward desired behaviors, ensuring they *"do the right thing"* in increasingly complex and ethically-charged scenarios.

Traditional control methods such as MPC, or LQR in robotics often rely on pre-programmed behaviors and lack the adaptability and learning capabilities of RL. However, RL lacks inherent performance guarantees in agent-environment interactions, requiring additional assurance measures [16]. Reward engineering provides a powerful tool for bridging the gap between conventional control and intelligent, learning-based systems. While early AI research could focus solely on achieving goals.

Therefore, this work aims to provide a comprehensive review of reward design, investigating the key concepts,

trends, challenges, and opportunities in reward shaping/engineering. It explores how the design of reward functions plays a crucial role in influencing the behavior of learning agents in various tasks and environments. Moreover, it investigates how advancements in reward engineering are paving the way for more efficient and effective learning algorithms. This review sheds light on the importance of reward design in RL and highlights the potential of reward engineering to drive innovation in the fields of RL and Artificial Intelligence (AI). As AI agents become more sophisticated and autonomous, the design of reward mechanisms becomes a crucial, yet increasingly challenging aspect of their development.

This survey provides the first comprehensive exploration of reward design in RL, addressing a significant gap in the literature. While a chapter in [3] briefly discusses the fundamentals of reward function design and its connections to behavioral sciences and evolution, it does not offer the level of analysis presented in this work. We present a detailed taxonomy of reward shaping techniques, examining their advantages, limitations, and application domains. In contrast to prior studies with narrower scopes, this review integrates a broader range of methodologies, offering a more comprehensive perspective on reward shaping and engineering.

The motivations for this review stem from the increasing complexity of modern AI systems, which necessitates a nuanced approach to reward design. It highlights the crucial role of well-crafted reward functions in guiding intelligent agents toward desired behaviors, emphasizing the balance between informativeness and sparsity in reward signals. Moreover, the review underscores the practical applications of reward design in various domains, particularly in robotics. It explores the implications of reward engineering for bridging the gap between simulated and real-world scenarios.

By investigating both the theoretical foundations and real-world applications of reward design, this review aims to facilitate a deeper understanding of how effective reward mechanisms can enhance learning efficiency. It identifies current challenges and opportunities for innovation, encouraging future research to advance the field of RL and artificial intelligence.

Hence, our work is structured as follows: Section I presents this review with an introduction. Section II details the process of selecting papers. Section III provides the background and fundamental concepts. Section IV outlines the taxonomy of reward shaping and engineering. Section V illustrates the real-world applications in robotics and other domains. A comprehensive exploration of Sim-to-Real is presented in Section VI. Section VII discusses the shortcomings and advantages of reward engineering. Finally, the open challenges and future directions, along with the conclusion, are presented in Sections VIII and IX, respectively.

## II. PAPERS SELECTION PROCESS
A comprehensive literature review was conducted using a systematic search strategy across multiple databases.

To ensure the transparency and reproducibility of our findings, we adhered to the PRISMA 2020 guidelines [17] for conducting and reporting systematic literature reviews. The search focused on reward shaping techniques in AI-based control systems, yielding a final set of 55 relevant papers after rigorous filtering and manual review. To ensure the reliability of the review, a quality assessment was performed on each included study using a set of benchmark questions.

Search terms included combinations of ''reward shaping,'' ''reward engineering,'' ''reinforcement learning,'' ''reward design,'' ''machine learning,'' and ''control systems.'' Due to the recentness of this topic, there is a limited number of papers that we potentially can review, therefor our inclusion criteria were established to focus on studies published in English between 1999 and 2024, which directly addressed the application of reward shaping techniques in AI-based control systems, and presented empirical results. Papers solely focused on theoretical frameworks or lacking empirical data were excluded, except for [6], [18], [19], and [20] because of their high impact on our review.

## III. BACKGROUND AND FUNDAMENTAL CONCEPTS
Reward shaping becomes more advantageous as the likelihood of an agent wasting time exploring pointless areas of the environment increases, and thus well-designed reward shaping can more effectively guide exploration, in addition to reward shaping, reward design has also been explored as a means of policy teaching, [21], [22], [23]

To define Reward Shaping and Reward Engineering, it is essential to go through the core RL concepts:

- Agent: An entity that interacts with the environment.
- Environment: The agent's world, governed by specific rules and dynamics.
- Policy: A function mapping states to actions, defining the agent's behavior.
- Reward: A signal received by the agent based on its actions, indicating the desirability of a state or action.
- Value Function: A function that estimates the expected future reward for a given state or state-action pair.
- Exploration vs. Exploitation: The balance between trying new actions and exploiting known good ones.

Based on the findings [6], it can be stated that **Reward shaping:** the most common name for reward engineering, shaping or design methods, is a technique inspired by animal training where supplemental rewards are provided to make a problem easier to learn. This includes modifying the original reward function by introducing additional incentives or penalties to guide the agent's learning process. **Reward engineering:** encompasses a broader set of techniques, including using other algorithms to design the reward function or designing reward functions from scratch.

The concept of reward plays a pivotal role in the field of artificial intelligence, particularly within the framework of reinforcement learning.

Two distinct perspectives on the nature and sufficiency of reward have emerged, sparking debate among researchers

The first perspective, championed by the "Reward is Enough" hypothesis [18], posits that maximizing a scalar reward, a single numerical value representing progress towards a goal could be the key to understanding and further building artificial intelligence. This view proposes that complex cognitive abilities, like learning, language, and social intelligence, emerge as a consequence of striving for this reward. The complexity of the environment, it is argued, naturally drives agents to develop these abilities to achieve their goals more efficiently. However, the "Scalar Reward is Not Enough" perspective challenges this view [19], asserting that relying solely on a single numerical value fails to capture the multi-faceted nature of human intelligence, particularly in domains involving ethical considerations or complex, subjective goals. This perspective advocates for the use of vector-valued rewards, which represent multiple aspects of progress and can better guide the development of safe, human-aligned AI. While both perspectives acknowledge the importance of reward in learning, the debate centers on the sufficiency of a single scalar value in representing the full spectrum of intelligent behavior. The "Scalar Reward is Not Enough" perspective highlights the need for more nuanced reward representations that can account for the intricate and often context-dependent nature of human intelligence.

### A. GENERAL PITFALLS IN REWARD DESIGN

As was established, reward design can be challenging and time-consuming, furthermore, its effects usually can be noticeable in the behavior of the agent in addition to the environment.

For a given task, understanding the difficulties in reward engineering can help to determine the most compatible, suitable, and successful reward shaping techniques:

- Reward Sparsity: Lack or delay of frequent reward signals can lead to slow learning.
- Deceptive Rewards: Reward signals may encourage the agent to find "easy" solutions that are not aligned with the true objective.
- Reward Hacking: Agents may exploit unintended loopholes in the reward function to achieve high rewards without fulfilling the desired goal.
- Unintended Consequences: Reward designs can lead to unexpected and undesirable behaviors due to the complex interplay between agent actions and the environment.
- Misaligned Reward with True Objective: This highlights the crucial problem of ensuring the reward function actually incentivizes the desired behavior.
- Reward Function Complexity: A complex reward function with multiple factors can be difficult to design and interpret.
- Difficulty in Evaluating Reward Design: It can be difficult to objectively evaluate the effectiveness of a reward function, especially in complex environments.

These are some of the factors why Reward Engineering could be an acceptable substitute for conventional reward design.

### B. SCALAR VS VECTOR REWARD

In RL, the choice between scalar and vector rewards significantly impacts the agent's learning process. Scalar rewards, represented by a single numerical value, provide a simple measure of progress towards a goal. This simplicity makes them computationally efficient and easy to interpret. However, they often fail to capture the nuances of complex tasks, potentially leading to suboptimal behavior. Vector rewards, on the other hand, utilize multiple values to represent different aspects of the task, offering a more comprehensive evaluation of the agent's actions. This richer feedback allows for a finer-grained learning process, enabling the agent to prioritize specific aspects of the task and potentially achieve more desirable outcomes. While vector rewards can lead to improved performance, they come with the challenge of designing effective reward functions, balancing multiple objectives, and navigating the computational complexities associated with multi-dimensional feedback. The choice between scalar and vector rewards depends on the specific task, the desired level of performance, and the available computational resources.

## IV. TAXONOMY OF REWARD SHAPING/ENGINEERING TECHNIQUES

This section contains a categorization of reward shaping techniques based on underlying principles, along with detailed explanations of each category, specific algorithms, advantages, disadvantages, and evaluation criteria.

Psychologists differentiate between extrinsic motivation, where actions are driven by specific anticipated rewards, and intrinsic motivation, where actions are motivated by inherent enjoyment [24]. Similarly in RL, Rewards can be categorized into intrinsic and extrinsic types, illustrated in Figure 1.

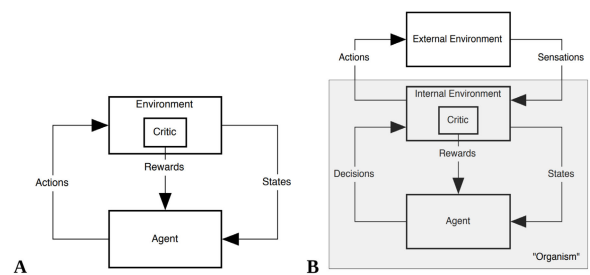

**FIGURE 1.** Agent-environment interaction in reinforcement learning. A: The agent receives rewards from a "critic" within its environment. B: This panel expands on Panel A by distinguishing between an internal and an external environment, with rewards originating from the internal environment. The shaded box represents what could be likened to the "organism." [Figure 1] [24].

These rewards play critical roles in motivating and guiding agent behavior, making the shaping and design of these functions crucial for determining the success or failure of algorithms. In their study of reward functions [25] introduced a computational framework aimed at optimizing these rewards to improve agent behaviors. Their approach focuses on defining optimal reward structures based on fitness functions and environmental distributions. Similarly, [26] further explored intrinsic motivation within the RL framework. They proposed an evolutionary-inspired optimal reward framework that emphasizes designing reward functions to enhance evolutionary success across diverse environments. This framework explores the delicate balance between intrinsic and extrinsic motivations, emphasizing the computational modeling of intrinsic motivation driven by internal rewards crucial for intellectual growth. They define the optimal reward as:

$$r_{\mathcal{A}}^* = \arg\max_{r_{\mathcal{A}} \in \mathcal{R}_{\mathcal{A}}} \mathbb{E}_{E \sim P(\varepsilon)} \mathbb{E}_{h \sim \langle A(r_A), E \rangle} \{\mathcal{F}(h)\}, \quad (6)$$

where $\mathbb{E}$ denotes the expectation operator. A special reward function in $\mathcal{R}_{\mathcal{A}}$ is the fitness-based reward function, that most directly translates fitness $\mathcal{F}$ into an RL reward function, i.e., where the fitness function $\mathcal{F}$ measures a scalar evaluation of the agent based on its interaction history $h$, which is sampled from the distribution resulting from its interaction with the environment. The fitness value of a lifetime-length history is the cumulative fitness-based reward for that history, [26, Equation 1].

Previous studies highlight the importance of advanced reward shaping methodologies in improving agent learning and adaptation.

### A. POLICY GRADIENT METHODS

Policy gradient methods directly optimize an agent's policy to maximize cumulative rewards [27]. Unlike value-based approaches, which estimate value functions, policy gradient methods focus solely on improving the policy itself.

#### 1) POLICY GRADIENT FOR REWARD DESIGN (PGRD)

In RL, an agent aims to maximize its total reward over time, referred to as its return. It's crucial to note that the sequence of environment states is influenced by the selected reward function, thereby affecting the goals of the agent's designer. The challenge of finding the optimal reward stems from the fact that, while the objective reward function is fixed in the problem formulation, the choice of reward function remains within the designer's control [28]. Addressing this challenge, previous research introduced the PGRD, which employs online gradient ascent to iteratively adjust reward parameters during the agent's operation. They argue that "the optimal reward parameters are determined by solving the optimal reward problem" [28, Equation 1]:

$$\theta^* = \arg\max_{\theta \in \Theta} \lim_{N \to \infty} \mathbb{E}\left[\frac{1}{N}\sum_{t=0}^{N} \mathcal{R}_{\mathcal{O}}(s_t)|\mathcal{R}(\cdot, \theta)\right], \quad (7)$$

where $\mathcal{R}_{\mathcal{O}}$ is objective reward function given by the designer and $\mathcal{R}(\cdot)$ is separate agent reward function. Authors abstractly represent this selection by parameterizing the reward with a vector of parameters $\theta$ chosen from a parameter space $\Theta$. Each $\theta \in \Theta$ defines a reward function $R(\cdot; \theta)$, which subsequently generates a distribution over the sequences of environment states using the agent's RL method. The expected return achieved by the designer for the choice of $\theta$ is denoted as $J(\theta)$.

PGRD adjusts to accommodate the agent's abilities and uncertainties in model accuracy. It formalizes updates to the reward function and policy by estimating gradients of the objective return and policy, ensuring convergence with rigorous proofs. Results illustrate its effectiveness in discrete-time, partially observable environments, as depicted in Figure 2, to maximize the expected mean objective reward over an infinite horizon. Unlike traditional methods that impose fixed goals on agents, PGRD enhances results compared to conventional policy gradient methods, highlighting its capacity to improve agent performance across various scenarios through dynamic reward optimization.

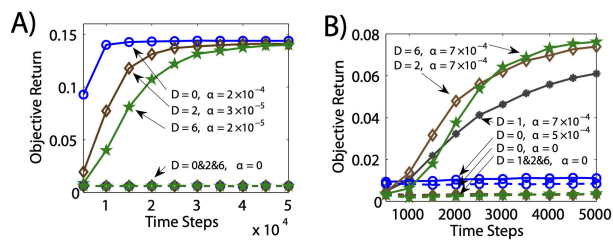

**FIGURE 2.** PGRD performance with A) poor model, B partially observable world, [28, Figure 3].

#### 2) LEARNING INTRINSIC REWARD FOR POLICY GRADIENT (LIRPG)

In a similar vein, [29] introduced the LIRPG algorithm. LIRPG enhances RL by enabling agents to dynamically learn intrinsic rewards in addition to traditional extrinsic rewards. They propose the following equation for updating policy parameters $\theta$ by incorporating both policy and intrinsic reward parameters through regular policy gradient updates [29, Equation 4]:

$$\theta' \approx \theta + \alpha G^{ex+in}(s_t, a_t)\nabla_\theta \log \pi_\theta(a_t|s_t). \quad (8)$$

LIRPG has demonstrated improved learning efficiency and performance in complex environments, reducing sample complexity and accelerating learning processes. Results indicate that LIRPG optimizes agent performance across Atari games and Mujoco domains, as illustrated for Hopper and HalfCheetah in Figure 3, consistently outperforming baseline agents that rely solely on extrinsic rewards. However, challenges include the necessity for meticulous parameter tuning and potential sensitivity to the selection and interaction of intrinsic reward functions with policy updates. It is noteworthy that LIRPG expands on the concept of reward shaping by allowing agents to adjust their behavior based on internal signals of progress or success, alongside the external rewards provided by the environment.

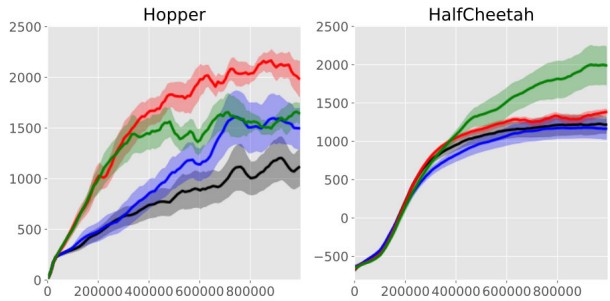

**FIGURE 3.** The x-axis represents time steps during the learning process, while the y-axis denotes the average reward over the last 100 training episodes. The black curves correspond to the baseline PPO architecture. The blue curves represent the PPO-live-bonus baseline. The red curves depict our augmented architecture using LIRPG. The green curves show the performance of our LIRPG architecture where the policy module was trained solely with intrinsic rewards. The darker curves represent averages across 10 runs with different random seeds. The shaded area indicates the standard errors across these 10 runs, [29], Figure 4].

### 3) DDPG FROM DEMONSTRATIONS (DDPGFD)

Deep Deterministic Policy Gradient (DDPG) is a model-free, off-policy reinforcement learning algorithm that combines the strengths of Deep Q-Learning (DQN) and Deterministic Policy Gradient (DPG). It is designed for environments with continuous action spaces, using an actor-critic approach to learn both a policy and a Q-function concurrently.

As an extension of the DDPG algorithm [30], DDPGfD is tailored for robotic RL tasks with sparse rewards. DDPGfD employs off-policy learning by integrating demonstration trajectories into the replay buffer, thereby leveraging human-provided guidance to bootstrap learning and address exploration challenges common in high-dimensional control problems such as robotics. The algorithm modifies DDPG in several key ways: integrating transitions from a human demonstrator into the replay buffer and utilizing prioritized replay to effectively sample transitions from both demonstration and agent data. The learning process includes a mix of 1-step $L_1(\theta^Q)$ and n-step return $L_n(\theta^Q)$ losses to enhance performance. Furthermore, it updates multiple times per environment step, thereby improving learning efficiency. Regularization is implemented with L2 penalties on the critic's weights $L_{reg}^C(\theta^Q)$ and the actor's weights $L_{reg}^C(\theta^\pi)$, promoting stable training and generalization. The experimental setup involves tasks of inserting a two-pronged deformable plastic clip into a housing using a 7-DOF robotic arm, with results depicted in Figure 4.

DDPGfD simplifies learning by integrating human guidance and eliminates the need for intricate reward shaping. However, significant challenges persist in maintaining demonstration quality and scalability across a wide range of robotic applications.

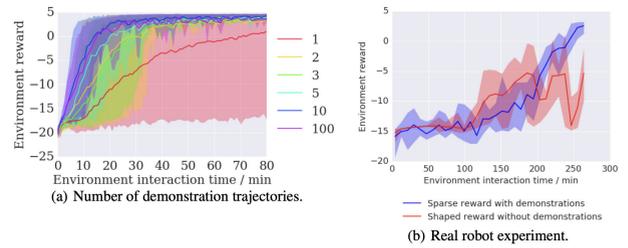

**FIGURE 4.** (a) Learning curves illustrating DDPG from Demonstrations (DDPGfD) performance on the clip insertion task with different amounts of demonstration data. DDPGfD demonstrates the ability to solve sparse-reward tasks effectively with minimal human demonstration, showcasing robust learning capabilities. (b) Results from 2 runs conducted on a physical robot. DDPGfD exhibits accelerated learning compared to DDPG and achieves performance without requiring handcrafted reward functions, as shown in [30], Figure 4].

### B. METHODS WITH ROBUSTNESS AND ADAPTABILITY

Robustness and adaptability are critical aspects of RL, robustness ensures RL agents perform well under uncertain or changing conditions by handling disturbances and model inaccuracies, while adaptability allows RL models to adjust to new environments, tasks, or dynamics, ensuring flexibility beyond initial training [31].

### 1) LEADER-FOLLOWER FRAMEWORK

Reward shaping offers a way to enhance robustness [31], [32]. A leader-follower framework employs a technique similar to that described in [33], where the leader, by modifying the follower's reward function, seeks to influence their actions in the desired direction, therefore enhancing the robustness of the system.

Nevertheless, when dealing with real-world rewards collected from sensors, these sensors are often affected by noise, which can distort reward signals and result in sub-optimal performance in RL models. To tackle these issues, [32] introduces a robust RL framework that uses a confusion matrix to estimate and correct noisy rewards. This framework incorporates an unbiased estimation algorithm designed to function without making assumptions about the underlying distribution of errors. The method involves defining unbiased surrogate rewards $\hat{r}$ based on estimated confusion matrices [32], Theorem 1] and applies the Q-learning algorithm with surrogate rewards [32], Equation 3]:

$$\mathcal{Q}_{t+a}(s_t, a_t) = (1 - \alpha_t)\mathcal{Q}(s_t, a_t) + \alpha_t \left[ \hat{r}_t + \gamma \max_{b \in \mathcal{A}} \mathcal{Q}(s_{t+1}, b) \right], \quad (9)$$

where $\alpha \in (0, 1)$ : the learning rate, will converge to the optimal $Q - function$ as long as $\sum_t \alpha_t = \infty$ and $\sum_t \alpha_t^2 < \infty$. Experimental validation on platforms such as OpenAI Gym and Atari games demonstrate notable performance improvements for trained policies, especially when combined with the Proximal Policy Optimization (PPO) algorithm. This integration results in higher expected rewards under noisy conditions, as shown in Figure 5. Additionally, the framework sometimes achieves greater cumulative rewards by utilizing noise for better exploration and employing noise-removal mechanisms.

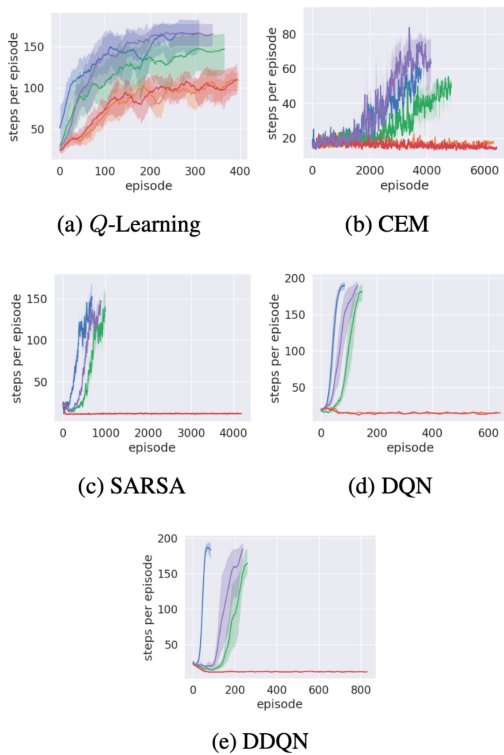

(a) *Q*-Learning

(b) CEM

(c) SARSA

(d) DQN

(e) DDQN

**FIGURE 5.** Learning curves from five rewards robust RL algorithms on CartPole game with true rewards (*r*) ■, noisy rewards (*r̃*) ■, sample-mean noisy rewards ■, estimated surrogate rewards (*r̂*) ■, and sample-mean estimated surrogate rewards ■, [32, Figure 4].

However, the method's computational complexity may present challenges, especially in real-time or resource-limited scenarios. Additionally, its effectiveness across various noise patterns and environments hinges on precise noise characterization and parameter tuning. Ultimately, this process can be viewed as a form of reward engineering, aimed at adjusting or refining the reward function to mitigate noise impact and enhance learning process reliability.

Similarly, to ensure that reinforcement learning (RL) policies meet specific control criteria, [34] introduces a systematic approach for shaping rewards to align optimal policy trajectories with these requirements. This method tackles the challenge of achieving control objectives such as settling time and steady-state error without explicit models of system dynamics.

## C. EXPLORATION STRATEGIES
In RL, exploration strategies are techniques that balance the trade-off between exploration (discovering new information) and exploitation (leveraging existing knowledge) [35]. These strategies help RL agents learn optimal policies by exploring their environment while exploiting learned knowledge.

### 1) HASH-BASED REWARD SHAPING
The effectiveness of count-based exploration methods in small, discrete spaces contrasts with the challenges they face in larger, continuous environments where state re-encounters are infrequent. In their study on extending count-based exploration to high-dimensional and continuous state spaces, [36] introduced the use of hash codes to facilitate state counting and exploration. This method employs static hashing techniques, such as locality-sensitive hashing SimHash, which retrieves a binary code of state $s \in \mathcal{S}$ as described in [36, Equation 2]:

$$\phi(s) = sgn(Ag(s)) \in \{-1, 1\}^k, \qquad (10)$$

where g is an optional preprocessing function and A is a matrix with i.i.d. entries drawn from a standard Gaussian distribution $\mathcal{N}(0, 1)$. Additionally, the study uses learned hashing via autoencoders (AE) to assign exploration bonuses based on state visitation counts, as illustrated in Figure 6.

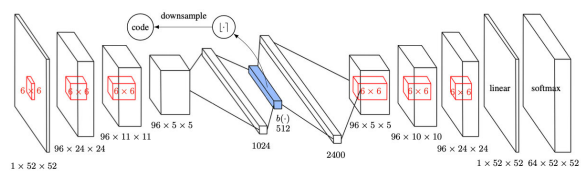

**FIGURE 6.** The architecture of the AE, [36, Figure 1].

### 2) VARIATIONAL INFORMATION MAXIMIZING EXPLORATION (VIME)
VIME is an exploration strategy designed for continuous control tasks, focusing on maximizing the agent's understanding of the environment's dynamics. It employs variational inference within Bayesian neural networks, enabling efficient handling of complex state and action spaces. VIME encourages exploration by using information gained from the learned dynamics model as intrinsic rewards. This motivates the agent to seek both external rewards and novel experiences.

Empirical evidence (see Figure 7) shows that VIME outperforms heuristic exploration methods across various continuous control tasks, including those with sparse rewards. However, the reliance on Bayesian neural networks for dynamics modeling introduces computational complexity. Future research should explore alternative methods for quantifying surprise and using the learned dynamics model for planning. VIME's adaptability to high-dimensional spaces and potential applicability across different reinforcement learning domains are strengths. However, challenges remain in tuning hash functions and ensuring robust performance across environments due to sensitivity to the quality of state representations.

The results illustrated in Figure 7 compare Trust Region Policy Optimization (TRPO) (baseline), TRPO-SimHash, and Variational Information Maximizing Exploration (VIME) [37] on tasks such as MountainCar, Cart-PoleSwingup, HalfCheetah, and SwimmerGather.

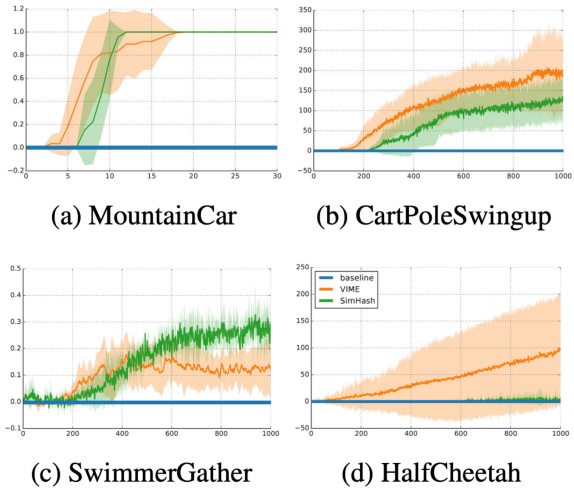

(a) MountainCar

(b) CartPoleSwingup

(c) SwimmerGather

(d) HalfCheetah

**FIGURE 7.** The mean average return of different algorithms on rllab [38] tasks with sparse rewards. The solid line shows the mean average return, while the shaded area represents one standard deviation, over 5 seeds for the baseline and SimHash (with baseline curves overlapping the axis). Count-based exploration with hashing enables goal achievement in all environments (indicated by a nonzero return), while baseline TRPO with Gaussian control noise fails completely. TRPO-SimHash effectively captures the sparse reward on HalfCheetah but does not perform as well as VIME. SimHash's performance is comparable to VIME on MountainCar and surpasses VIME on SwimmerGather [36, Figure 3].

### 3) ONLINE REWARD SHAPING (EXPLORES)

To enhance the learning efficiency of reinforcement learning (RL) agents dealing with sparse or noisy reward signals, [39] introduce Exploration-Guided Reward Shaping (EXPLORS). This framework combines intrinsic reward learning with exploration-based bonuses in a fully self-supervised manner, aiming to maximize the agent's effectiveness in obtaining extrinsic rewards:

---

**Algorithm 1** Online Reward Shaping, [39, Algorithm 1]

1: **Input:** Extrinsic reward $\bar{R}$, and RL algorithm $L$
2: **Initialization:** $\pi_0, \hat{R}_0$
3: **for** $k = 1, 2, \ldots, K$ **do**
4:     update policy $\pi_k \leftarrow L(\pi_{k-1}, \hat{R}_{k-1})$
5:     update reward $\hat{R}_k$ using $\hat{R}_{k-1}$ and $\pi_k$
6: **end for**
7: **Output:** $\pi_K$

---

Experimental findings across various environments validate the method's effectiveness in accelerating learning compared to conventional RL approaches, demonstrating its capability to surpass the limitations of traditional reward shaping methods, as shown in Figure 8. This is particularly evident in scenarios where traditional methods are impractical or ineffective. However, the study also acknowledges several constraints, including the need for extensive evaluation in more complex environments and the challenges associated with effectively combining intrinsic rewards with exploration bonuses.

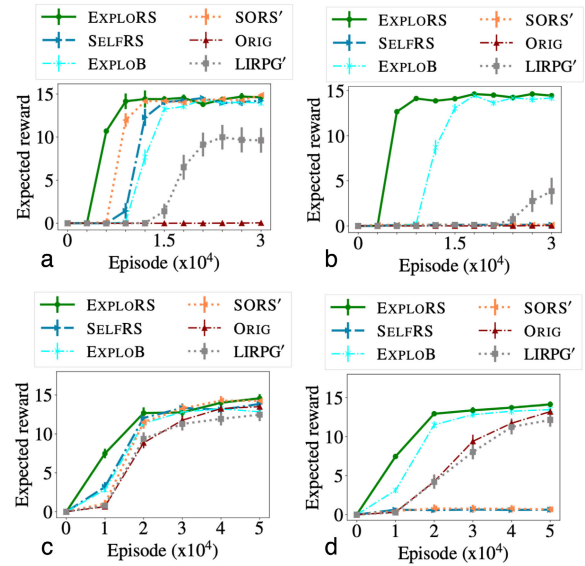

**FIGURE 8.** a) Room$^0$, REINFORCE b) Room$^+$, REINFORCE c) LineK$^0$, REINFORCE d) LineK$^+$, REINFORCE. These plots illustrate the agent's convergence in performance relative to training episodes. (a, b) display results for the REINFORCE agent on Room$^0$ (ROOM variant without any distractor state) and Room$^+$ (ROOM variant with a distractor state). (c, d) present results for the REINFORCE agent on LineK$^0$ (LineK variant without any distractor state) and LineK$^+$ (LineK variant with distractor states) [39, Figure 5].

Within the same framework, [40] introduce Reward Uncertainty for Exploration (RUNE), a method within preference-based RL algorithms that integrates uncertainty in learned reward functions as an exploration bonus. By incorporating the variance in predictions from an ensemble of reward functions optimized to align with human feedback, RUNE aims to enhance exploration in environments where learning is guided by human preferences, as depicted in Figure 9.

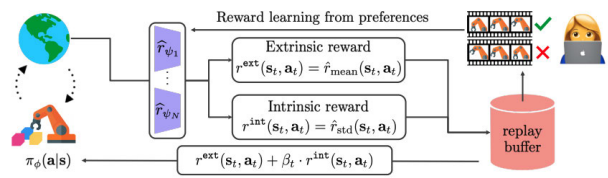

**FIGURE 9.** The agent interacts with the environment and learns an ensemble of reward functions based on teacher preferences. Each state-action pair's total reward combines the extrinsic reward with the mean of the ensemble's predicted values and the intrinsic reward with the standard deviation among the ensemble's predictions [40, Figure 1].

This approach offers several benefits: it provides a systematic way to balance exploration and exploitation by prioritizing actions with higher uncertainty in expected rewards, potentially accelerating learning and improving sample efficiency compared to traditional preference-based

RL methods. Experimental results in Figure 10 demonstrate RUNE's effectiveness in enhancing asymptotic success rates and overall feedback efficiency during training scenarios. However, challenges remain, such as developing robust techniques to manage varying levels of reward uncertainty and addressing the computational overhead of maintaining an ensemble of reward function predictions. Nevertheless, this work represents a significant advancement in RL research by proposing a mechanism to integrate reward uncertainty into exploration strategies, paving the way for future developments in adaptive learning algorithms driven by human feedback.

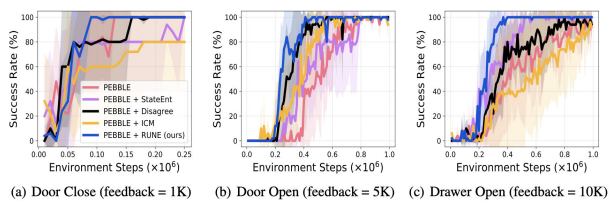

**FIGURE 10.** Learning curves on robotic manipulation tasks measuring success rates. Exploration methods consistently improve the sample efficiency of PEBBLE. Notably, RUNE shows larger gains compared to other existing exploration baselines. The solid line represents the mean and the shaded regions denote standard deviations across five runs [40, Figure 3].

### D. POLICY PARAMETERIZATION
Policy parameterization refers to explicitly modeling the policy as a function of learnable parameters (often denoted as $\theta$). These parameters determine the agent's behavior, and common choices include neural network weights [41]. By optimizing these parameters, we improve the policy's performance through gradient-based updates.

Continuous control tasks traditionally rely on complex neural network methods. However, [42] examines the effectiveness of simpler policy parameterizations, such as linear and radial Basis Function (RBF) policies, in these settings. The primary concept is to enrich the representational capacity by using random Fourier features of the observations. These features are defined as [42, Equation 6]:

$$y_t^{(i)} = \sin\left(\frac{\sum_j P_{ij} s_t^{(j)}}{v} + \phi^{(i)}\right), \qquad (11)$$

where each element of $P_{ij}$ is drawn from $\mathcal{N}(0, 1)$, $v$ is the bandwidth parameter, and $\phi$ is a random phase shift. The study shows that these streamlined policies can achieve competitive performance on benchmarks such as OpenAI Gym and can be trained faster than neural networks. Nevertheless, conventional training methods often result in policies susceptible to perturbations, which is critical for real-world applications. By diversifying initial state distributions during training, the research demonstrates enhanced policy robustness, similar to the benefits observed with model ensembles and domain randomization.

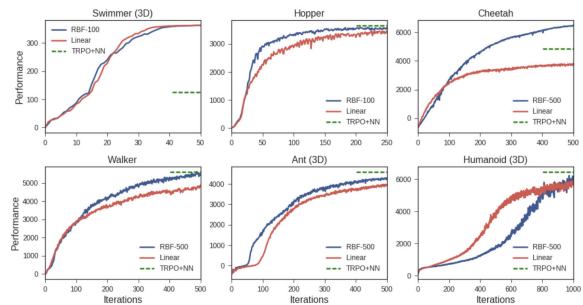

**FIGURE 11.** Learning curves for the Linear and RBF policy architectures. The green line indicates the reward achieved by neural network policies on the OpenAI Gym website as of 02/24/2017 (trained with TRPO), [42, Figure 1].

Despite these advantages, simple policy approaches may face challenges. Their scalability to extensive state and action spaces can be limited compared to the adaptability of neural networks. Sensitivity to initial state distributions necessitates careful tuning for robustness. In complex scenarios demanding intricate behaviors, simple policies may not match the peak performance achievable with advanced neural networks. Additionally, their rigidity may hinder adaptation to dynamic or intricate environments compared to more flexible neural network architectures.

### E. INVERSE REWARD DESIGN
Inverse Reward Design (IRD) tackles the challenge of inferring the true objective behind a designed reward function. Instead of manually engineering rewards, IRD allows RL agents to learn from expert demonstrations and deduce the underlying intention driving those actions.

In RL, the agent selects an action in a known state and receives a reward generated by a reward function $R$ that may be unknown to the agent The state transitions are based on the previous state and action, which is described by the transition function $T$, which may also be unknown. Conversely, in Inverse Reinforcement Learning (IRL), the inputs and outputs for the learner $L$ are reversed. $L$ observes the states and actions $\{(s, a), (s, a), \ldots, (s, a)\}$ of an expert $E$ (or its policy $\pi_E$), and learns a reward function $\hat{R}_E$ that best explains $E$'s behavior as the output [43].

IRL emphasizes the inference of the reward function that an expert agent is presumed to be maximizing based on observed behavior. This approach provides a robust framework for understanding and designing reward structures in various RL applications. This paper aims to underscore the significance of IRL and to enhance the discourse by providing an overview of the primary categories of IRL methodologies [43], [44], [45], [46], [47], [48].

The primary categories of IRL methods are classified as follows:

1) **Model-Based IRL**: In this approach, the reward function is inferred alongside a model of the environment's dynamics. The agent learns both the reward structure and the transition probabilities $T(s'|s, a)$. A common model-based method is Maximum Entropy IRL, which seeks to maximize the entropy of the policy while explaining the observed behavior.

2) **Model-Free IRL**: These methods directly infer the reward structure from demonstrations without explicitly modeling the environment. For instance, algorithms like Fitted IRL and Bayesian IRL estimate the reward function based solely on observed state-action pairs. The objective is to minimize the difference between the expected return of the inferred policy and that of the expert.

3) **Deep Inverse Reinforcement Learning**: Recent advancements in deep learning have led to the development of deep IRL techniques, which utilize neural networks to model complex reward functions. These methods, such as deep maximum entropy IRL, allow for the representation of intricate behaviors and expand the applicability of IRL in real-world scenarios. The learned reward function can be expressed as:

$$\hat{R}(s, a) = f_\theta(s, a), \quad (12)$$

where $f_\theta$ is a neural network parameterized by $\theta$.

It is important to note that the learned reward function may not exactly correspond to the true reward function, as illustrated in Figure 12.

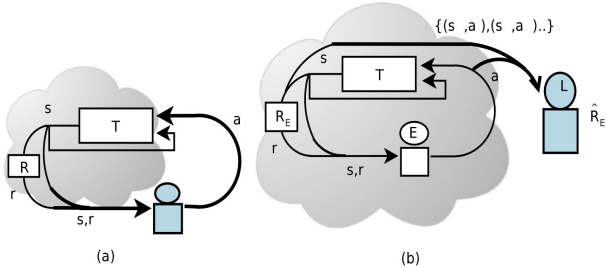

**FIGURE 12.** The figure depicts RL and IRL. The subject agent (shaded in blue), [43, Figure 3].

Coinciding with previous work, [49] proposed Inverse Reward Design (IRD), an approach that approximates the true reward function by treating the proxy reward as expert demonstrations. This method aids in planning risk-averse behavior and addresses problems like misspecified rewards and reward manipulation. IRD prevents harmful behaviors by avoiding dangerous areas, as shown in experiments with robots avoiding lava Figure 13. It also mitigates reward manipulation by treating designed rewards as observations rather than fixed objectives. IRD's risk-averse planning ensures agents avoid both known hazards and potential risks, enhancing safety and reliability. Additionally, IRD systems are robust to changes in high-dimensional feature spaces. However, IRD faces computational challenges in solving planning problems during inference and often relies on simplified assumptions about reward functions and environments, limiting its applicability in complex scenarios.

Accurate inference remains challenging, potentially leading to sub-optimal behaviors.

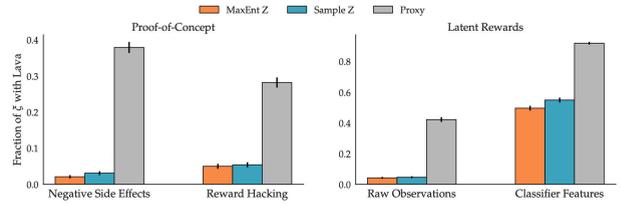

**FIGURE 13.** The results comparing [49] method to a baseline that directly plans with the proxy reward function, [49, Figure 4].

### F. REWARD HORIZON

The impact of using shaping to reduce the reward horizon is illustrated through a straightforward algorithm, Algorithm 2, that guarantees learning time is polynomial relative to the size of the critical region and independent of the MDP's size [50].

---

**Algorithm 2** The Horizon_Learn, [50, Figure 1]

---
1: **Input:** MDP $M$ and Reward Horizon $H$
2: **Initialization:** $\pi$
3: **while** successive episodes occur visiting only known states **do**
4:     Assign $s$ as the current state of $M$
5:     **if** $s$ is terminal **then**
6:         reset to $s_0$
7:     **end if**
8:     **if** $s$ is known **then**
9:         execute $\pi(s)$
10:     **else**
11:         execute any policy in $m_H(s)$ that still needs to be explored
12:         **if** $s$ becomes known **then**
13:             Select the action for $s$ in the optimal policy of $m_H(s)$, $a$
14:             Set $\pi(s) = a$
15:         **end if**
16:     **end if**
17: **end while**
18: **return** $\pi$

---

By shortening the reward horizon, they identify easier-to-learn MDPs, indicating that standard RL algorithms, such as Q-learning with $\varepsilon$-greedy exploration, can greatly benefit from this technique. Experimental data support that reducing the reward horizon speeds up learning, with shaping strategies leading to faster performance improvements compared to no shaping, as shown in Figure 14. Specifically, the time required to reach a certain performance level is significantly reduced with shaping: algorithms that explore based on reward feedback are notably quicker when the reward horizon is minimized.

The study emphasizes the crucial role of the reward horizon in determining the complexity of learning tasks,

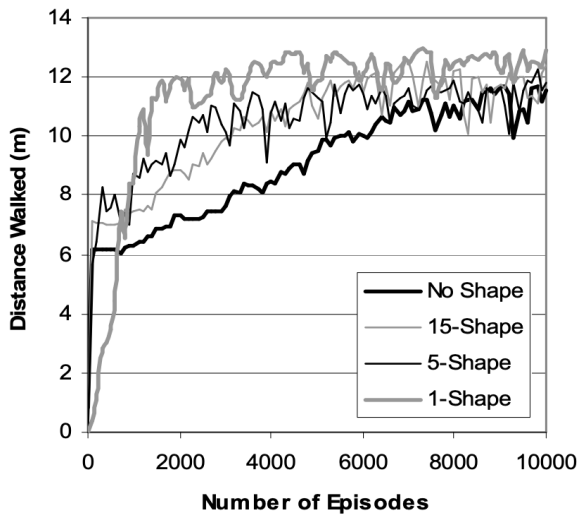

**FIGURE 14.** Average learning curves as the reward horizon is reduced from no shaping to shaping every action, [50], Figure 2].

showing that shaping can dramatically decrease the time required for RL agents to learn optimal policies. However, accurately determining and implementing the appropriate reward horizon is challenging, as improper shaping might not yield the expected acceleration in learning.

### G. POTENTIAL BASED METHODS
Potential-based methods [11] in RL focus on shaping the value function to guide an agent's behavior. By introducing auxiliary potentials, these methods encourage desired states and actions, leading to improved exploration and convergence. These methods usually involve defining a potential function, $\phi(s)$, over the state space, which captures the agent's desired progress towards a goal state.

$$R'(s, a) = R(s, a) + \gamma[\phi(s') - \phi(s)].$$ (13)

Guiding RL agents toward goals can be improved by analyzing cumulative rewards from episodes [51]. By reinforcing reward signals based on episode rewards, they proposed the Potential-Based Reward Shaping (PBRS) method to enhance learning efficiency and performance in both single-task and multi-task environments within the Arcade learning domain Figure 15. The reward function must also handle sparse reward signals by analyzing agent transitions in environments with sparse rewards. The method proposed the following function [51], Equation 12]:

$$\phi(s, a, t) = \begin{cases} 0 & \mathcal{R}(s, a) = 0 \\ 1 + \dfrac{\mathcal{R}^{ep} - \mathcal{R}_u^{ep}(t)}{\mathcal{R}_u^{ep}(t) - \mathcal{R}_l^{ep}(t)} & O.W, \end{cases}$$ (14)

where $\mathcal{R}(s, a)$ is the immediate reward, $\mathcal{R}^{ep}$ is the sum of rewards in the current episode (episode reward), $\mathcal{R}_u^{ep}(t)$ is the minimum value of episode reward until now, and $\mathcal{R}_l^{ep}(t)$ is the maximum value of episode reward until now. The potential function discourages unproductive states, reinforces positive rewards, and indicates the significance of negative rewards. This dynamic adjustment ensures continuous improvement and efficient exploration.

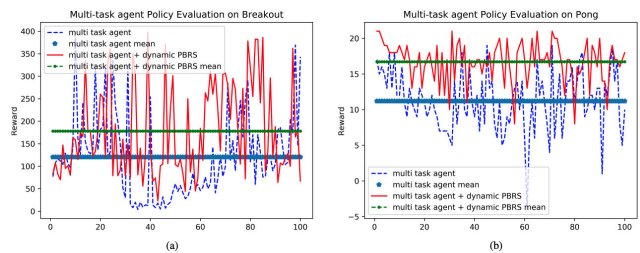

**FIGURE 15.** Results of PBRS on Pong and Breakout, [51], Figure 5].

Evaluations during learning and final policy assessments show that this method competes well with baseline methods, particularly in multitasking scenarios, thus advancing techniques for speeding up RL algorithms and improving adaptability in complex environments.

#### 1) THEORETICAL FOUNDATIONS FOR MULTI-AGENT SYSTEMS (MAS)
To investigate the theoretical implications of potential-based reward shaping in multi-agent systems, this research [20] extends previous findings, demonstrating that this technique maintains equivalence to Q-table initialization and does not affect the Nash Equilibria of the underlying stochastic game. Crucially, the study empirically reveals that potential-based shaping influences exploration, potentially leading to different converged joint policies. While the research focuses on fully observable domains, it highlights the potential of potential-based reward shaping for incorporating heuristic knowledge into multi-agent learning. The authors suggest that this technique can increase the likelihood of converging to a higher global utility and reduce convergence time, potentially mitigating the risks associated with unintended cyclical policies.

#### 2) PBRS-MAXQ-0 METHOD
Potential Based Reward Shaping (PBRS) and MAXQ are two extensively utilized techniques in reinforcement learning. Harutyunyan et al. [52] introduced PBRS-MAXQ-0, a novel algorithm that merges these methods within a hierarchical reinforcement learning (HRL) framework. PBRS-MAXQ-0 seeks to integrate heuristics into HRL tasks both effectively and efficiently, offering theoretical convergence guarantees under specific conditions, independent of additional rewards applied. Evaluations underscore several benefits: with appropriate heuristics, PBRS-MAXQ-0 notably accelerates convergence relative to the standard MAXQ-0 algorithm and competes well with other advanced MAXQ-based methods. Importantly, even with misleading heuristics, PBRS-MAXQ-0 shows resilience, eventually achieving convergence after prolonged learning periods, as illustrated in Figure 16.

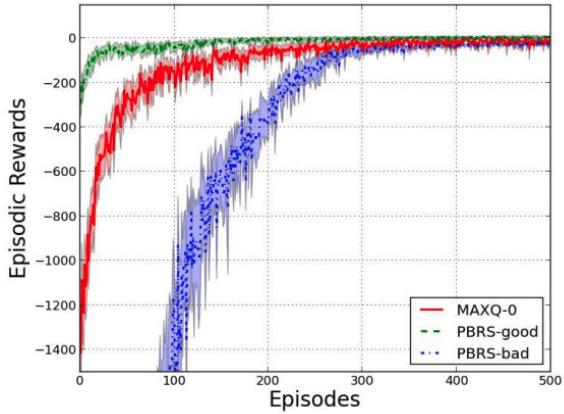

**FIGURE 16.** Performances of PBRS-MAXQ-0 in the Fickle Taxi problem, where PBRS-MAXQ-0 with reasonable heuristics (denoted by PBRS-good) and with misleading heuristics (denoted by PBRS-bad) [52, Figure 2].

However, integrating PBRS into MAXQ presents challenges, such as optimizing heuristic values and ensuring robust convergence across diverse environments. Fine-tuning PBRS parameters and maintaining heuristic accuracy are essential for maximizing the algorithm's efficacy across various HRL scenarios.

### 3) POTENTIAL-BASED ADVICE

Potential-based shaping functions ensure that policies learned with shaped rewards remain effective in the original MDP, maintaining near-optimal policies [11]. On the other hand, Reward shaping involves augmenting the reward function of MDP to accelerate learning by providing additional guidance beyond the intrinsic rewards of the MDP.

This paper [53] presents a novel framework for expressing arbitrary reward functions as potential-based advice within the context of reinforcement learning (RL). This method ensures policy invariance by learning an auxiliary value function, derived from a modified version of the original reward function. The potential-based approach facilitates efficient encoding of behavioral domain knowledge, leading to significant improvements in learning speed compared to existing methods. Notably, the framework introduces minimal computational overhead, as maintaining the auxiliary value function requires only linear time and space complexity.

While the proposed method demonstrates theoretical soundness, it currently lacks practical applications and specific algorithms. The paper primarily focuses on theoretical concepts, leaving further development of practical implementations for future research. The authors highlight the challenges associated with different reward shaping approaches: while potential-based methods offer guarantees, they often require extensive domain knowledge for effective implementation; auxiliary task-based methods offer flexibility but can become complex as the number of auxiliary tasks increases; and intrinsic motivation methods while promising for encouraging exploration, can be sensitive to parameter tuning.

The authors conclude that effective reward shaping requires a nuanced understanding of the specific problem domain and a careful balance between theoretical soundness and practical considerations. Future research should prioritize the development of more automated and robust reward shaping techniques to advance the field of RL.

### 4) PBRS IN EPISODIC REINFORCEMENT LEARNING

PBRS in episodic RL, explored by [54], sheds light on its applications in model-free, model-based algorithms, and multi-agent RL. It examines reward shaping in episodic tasks like games, revealing insights: potential-based shaping alters equilibria in stochastic games, introduces new equilibria with non-zero terminal state potentials, and reevaluates its role in PAC-MDP learning. It challenges the need for admissible potential functions, proving $\forall s \in Unknown$, $\Phi(s) \geq 0$ ensures optimistic exploration. The study provides analytical justification for PBRS, crucial in episodic RL with distinct initial and terminal states, emphasizing The potential role in enhancing learning efficiency.

### 5) CONSTRAINED RL WITH POTENTIAL-BASED REWARD FUNCTIONS (PBRF)

To generate safety-oriented aspects of reward functions from verified hybrid systems models [55] proposed an approach of using logically constrained RL to integrate formal methods and RL. Demonstrated on a standard RL environment for longitudinal vehicle control, this method showed faster convergence during training with augmented reward functions, particularly with logically constrained and potential-based reward functions (PBRF). The study found that partly auto-generated reward functions produced agents that generally maintained the safety level of hand-tuned reward functions, and reward scaling could emphasize certain aspects of the generated rewards. The training process was evaluated in terms of the number of epochs until convergence and the average accumulated reward across evaluation periods.

### 6) DIFFERENCE REWARDS INCORPORATING POTENTIAL-BASED REWARD SHAPING (DRIP) AND COUNTERFACTUALS AS POTENTIAL (CAP)

DRIP is introduced as a novel reward shaping technique for multi-agent reinforcement learning [56]. DRIP combines difference rewards, which incentivize agents to contribute to the overall system performance, with potential-based reward shaping. Potential-based shaping (see Equation 16) accelerates learning and maintains desired Nash equilibria by modifying the reward function using a potential function derived from domain-specific knowledge. While effective, DRiP requires a carefully designed potential function based on deep domain understanding, which can be challenging and time-consuming to obtain. CaP, on the other hand, automatically generates a dynamic potential function, eliminating the need for manual design and guaranteeing stable Nash equilibria. Although both CaP and DRiP capture similar knowledge, combining them doesn't provide extra benefits and can even negatively impact performance. For applications requiring theoretical guarantees, CaP is preferred; for performance priority, DRiP is recommended, especially when domain knowledge is readily available. The work [56] tested these techniques in various domains, showcasing DRiP's consistent outperformance in accelerating learning and achieving better policies.

### H. DYNAMIC POTENTIAL-BASED REWARD SHAPING (DPBRS)

For improving reinforcement learning in single and multi-agent systems, this method introduces a dynamic potential-based function to perform the reward shaping, and to guide agents without affecting the optimal policy [57]. This function changes over time, adapting to the current state of the agents and the environment. The idea can be represented as follows through the formula of Q-Learning:

$$Q(s, a) \leftarrow Q(s, a) + \alpha[r + F(s, s') + \gamma \max_{a'} Q(s', a') - Q(s, a)], \quad (15)$$

where $F(s, s')$ is the general form of any state-based shaping reward:

$$F(s, s') = \gamma \Phi(s') - \Phi(s). \quad (16)$$

For the dynamic potential based, we can extend Equation 16 and include $t$; the agent's arrival time in the previous state $s$, and $t'$ is its arrival time at the current state $s'$ (i.e., $t < t'$):

$$F(s, t, s', t') = \gamma \Phi(s', t') - \Phi(s, t). \quad (17)$$

The DPBRS maintains existing guarantees, its advantages include its ability to improve decentralized multi-agent learning through carefully designed, domain-specific potential functions. These functions foster cooperation by addressing challenges like coordination, information asymmetry, and scalability. While the potential benefits of this approach include faster convergence, enhanced cooperation, and greater robustness, the method lacks comparisons with other algorithms and lacks examples in robotics or highly non-linear systems, where it was tested on a 2-D maze for Single-Agent learning as we see in Figure 17, and on Boutilier's Coordination Game for Multi-Agent learning.

### I. UPPER CONFIDENCE BOUND VALUE ITERATION (UCBVI)

Well-designed reward shaping can significantly improve sample complexity and enhance exploration, leading to better performance compared to uninformed exploration strategies. This is demonstrated in the following work, which investigates the benefits of reward shaping in reinforcement learning (RL) through a combination of theoretical analysis and experimental validation.

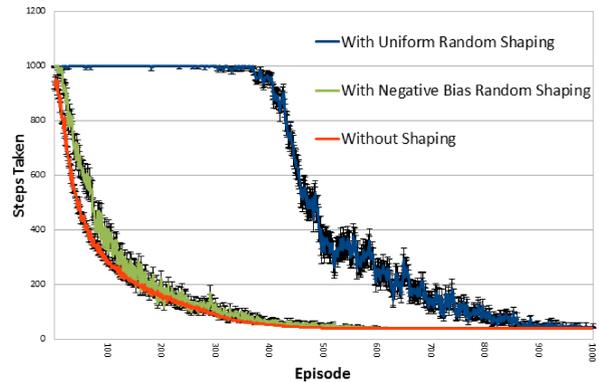

**FIGURE 17.** DPBRS single-agent maze results [57, Figure 2].

#### 1) UCBVI

The UCBVI algorithm [58] is a method for shaping rewards that extends the Value Iteration technique, ensuring that the resulting value function serves as an upper confidence bound (UCB) with a high probability on the optimal value function, It is important to note that this algorithm bears similarity to model-based interval estimation (MBIE-EB) [59]. The authors present proofs for their theorems and introduce several modified versions, including UCBVI-CH, which incorporates Chernoff Hoeffding's concentration inequality. However, no empirical results are provided.

#### 2) UCBVI-SHAPED

A modified version of the UCBVI algorithm [58] incorporates reward shaping to modify bonuses and value function projection [60]. Their analysis reveals that reward shaping can effectively prune irrelevant parts of the state space, sharpening optimism in a task-directed manner. This reduction in state space dependence leads to improved regret bounds and sample complexity benefits while retaining asymptotic performance.

The research was conducted on maze environments, comparing the performance of UCBVI-shaped with the standard UCBVI algorithm. Their findings show that reward shaping can enhance the performance of UCBVI, particularly in environments where agents are prone to wasting time exploring irrelevant areas.

This work contributes to a deeper understanding of reward shaping's impact on sample complexity and its potential to guide exploration more efficiently. It encourages future research to incorporate reward shaping more formally into sample complexity analysis, moving away from reward-agnostic approaches.

### J. DIFFERENCE REWARDS (D)

For single agent systems, difference rewards enhance the original reward function by incorporating a difference term that measures the discrepancy between the agent's current state and a designated reference state. This difference term serves as a guiding force, encouraging the agent to transition towards the reference state, which could be a desired goal state or a state representing a desirable condition. The modified reward function takes the form:

$$R'(s, a) = R(s, a) + \gamma[D(s', r) - D(s, r)], \quad (18)$$

where $R(s, a)$ is the original reward for taking action $a$ in state $s$, $R'(s, a)$ is the modified reward, $\gamma$ is a discount factor, $D(s, r)$ is the difference function, which measures the difference between the current state $s$ and the reference state $r$, and $s'$ is the next state after taking action $a$.

The simplicity of difference-based reward shaping makes it adaptable to various tasks and environments. However, its effectiveness hinges on a carefully chosen reference state, and in complex scenarios, it may not provide as comprehensive information as alternative methods.

For Multi Agents Systems (MAS) We define Difference Rewards $Di$ as [61], and [62] defined it:

$$Di(s_i, a_i) = G(s, a) - G(s - i \cup s_{c_i}, a - i \cup a_{c_i}). \quad (19)$$

The global system utility, denoted as $(G(s, a))$, depends on the system state $(s)$ and the joint action $(a)$. A counterfactual term, $(G(s - i \cup sc_i, a - i \cup ac_i))$, estimates the global utility without agent $(i)$'s contribution, considering states and actions excluding agent $(i)$ and fixed states and actions independent of agent $(i)$. The main idea of $Di$ in MAS is to encourage agents to contribute to the overall system utility by providing a reward that reflects the difference between the system's performance with and without the agent's contribution.

### 1) INDIVIDUAL AND DIFFERENCE REWARDS

RL can be used to find this optimal strategy in route choice problems in road networks, and we can test two different kinds of rewards. Individual rewards focus on the individual agent's benefit, leading to a selfish approach that can exacerbate congestion. Difference rewards $D$, on the other hand, promote system-wide optimization, leading to a more efficient and equitable allocation of traffic.

This work [63] explores two reinforcement learning approaches, IQ-learning and DQ-learning, for solving the route choice problem in road networks. Both use the same configuration but differ in their reward function. IQ-learning uses individual utility as the reward function, while DQ-learning uses a difference reward function that aims to maximize system utility. The difference reward function incentivizes agents to choose routes that minimize overall travel time, leading to a more balanced distribution of traffic across the network. Experiments show that DQ-learning significantly improves travel time compared to IQ-learning, successive averages, incremental assignment, and all-or-nothing assignment.

The experiment's findings, while significant, are specific to the scenario tested and may not generalize to other environments. It's important to note that both algorithms require an initial exploration phase to learn the environment effectively, and evaluating performance prematurely can lead to inaccurate conclusions. As we can see in the Figure 18, both algorithms tend to have similar convergence, but during the final episodes, DQ-learning shows better results. The study uses statistical testing with Gaussian distribution and confidence intervals to demonstrate the significance of DQ-learning's improvement. However, there are concerns about generalizing these results to other domains due to the specific focus on traffic assignment and the inherent assumptions about driver behavior in the model.

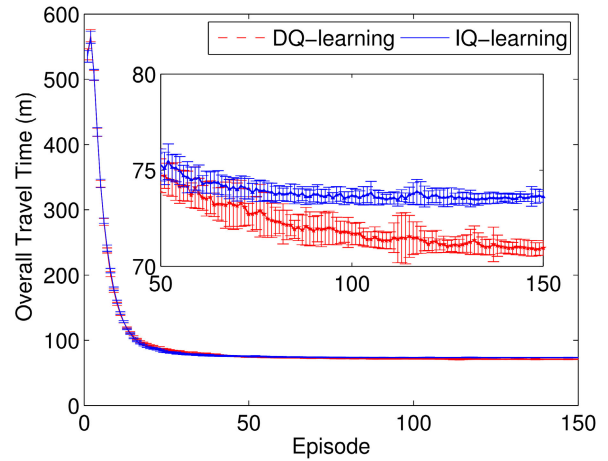

**FIGURE 18.** The main plot depicts convergence times and standard deviations across all episodes, while the inset focuses on the final episodes [63, Figure 3].

### K. KNOWLEDGE-BASED MULTI-OBJECTIVE MULTI-AGENT REINFORCEMENT LEARNING (MOMARL)

Reward shaping techniques Difference Rewards and potential-based reward shaping are popular methods for reward shaping. The research cited as [62] compares and evaluates these techniques in two studies, one using a novel benchmark problem called the Multi-Objective Beach Problem Domain (MOBPD) and the other using the Dynamic Economic Emissions Dispatch problem.

The results demonstrate that both D and PBRS can effectively guide agents towards Pareto optimal solutions in MOMARL domains, confirming that appropriate reward shaping is crucial in these settings. However, both techniques have limitations: D requires global knowledge of the system and the mathematical form of the evaluation function, while PBRS necessitates handcrafted potential functions which can be time-consuming and challenging to design effectively. Furthermore, the study found that D generally outperforms PBRS in terms of performance, especially when the required constraints are met.

The study concludes that the optimal technique for a given MOMARL application depends on specific constraints, such as the availability of system knowledge, bandwidth for communication, and the designer's expertise. The work establishes the MOBPD as a benchmark for future MOMARL research and highlights the need for further investigation into the design and application of these reward shaping methods.

### L. PLAN BASED METHODS

Plan-based methods combine the advantages of model-based planning and RL. These methods allow agents to simulate "what-if" scenarios and generate policy updates without causing state changes in the environment. Imagine it as agents using their imagination to explore different paths before taking real actions.

#### 1) UTILIZATION OF STRIPS

Plan-based reward shaping, as discussed by [64], utilizes domain knowledge to accelerate the convergence and improve the optimality of RL methods. It effectively addresses exploration challenges when integrated with model-free approaches. While model-based methods can mitigate these issues independently, the addition of reward shaping consistently boosts learning efficiency. The study introduces a STRIPS-based reward shaping technique, utilizing a Stanford Research Institute Problem Solver (STRIPS) representation of actions and goals through preconditions and effects. Compared to traditional MDP-based approaches, STRIPS-based shaping directs agents more effectively toward optimal policies. It suggests refining MDP-based reward shaping by focusing on the best path derived from the STRIPS plan rather than the entire state space's value function. Evaluations demonstrate the robustness of STRIPS-based shaping against plan knowledge errors, highlighting its superiority in enhancing policy quality and convergence speed. This approach provides a viable alternative to MDP-based methods, offering domain experts flexibility in expressing and utilizing domain knowledge.

#### 2) COMPARING PLAN BASED TO ABSTRACT MDP

Similarly [65] compared two reward shaping methods: plan-based and abstract MDP. The plan-based method supplements an agent's actions with additional rewards based on predefined plans. In contrast, the abstract MDP approach involves solving a higher-level MDP to shape behavior using its value function. The comparison evaluates these methods in terms of total reward, convergence speed, and scalability to complex environments. In large-scale settings, the plan-based method outperforms abstract MDP by offering detailed, sequential guidance to agents. However, in multi-agent scenarios with conflicting goals, abstract MDP excels due to its ability to manage coordination challenges better than plan-based methods. This highlights the importance of selecting reward shaping methods based on specific environmental characteristics to optimize agent performance effectively.

### M. BELIEF REWARD SHAPING (BRS)

Belief Reward Shaping (BRS) [66] enhances reinforcement learning by incorporating prior knowledge about the environment's reward structure. It augments the standard reward signal with "belief rewards" derived from a Bayesian framework, reflecting prior assumptions or beliefs about the distribution of rewards. This approach diverges from traditional methods by avoiding sole reliance on environmental interactions for learning the reward structure. BRS directly provides shaping rewards based on both the state and action, addressing a limitation of potential-based reward shaping (PBRS), which only considers the state.

By integrating these prior beliefs, the authors suggest that an agent's reward should be influenced not only by external environmental sensations but also by its internal belief system. This belief system, represented by an internal critic, is dynamically updated based on both prior beliefs and environmental sensations. The critic then provides an updated reward to the agent, taking into account both its prior beliefs and the current sensory information. BRS leverages Bayesian methods to specify prior beliefs on the environment's reward distribution, demonstrating that more complex and accurate prior beliefs lead to improved agent performance. Theoretical guarantees for BRS's consistency when augmenting Q-learning are provided, but these guarantees hold only if the true environment reward distribution falls within the critic's hypothesized set of models.

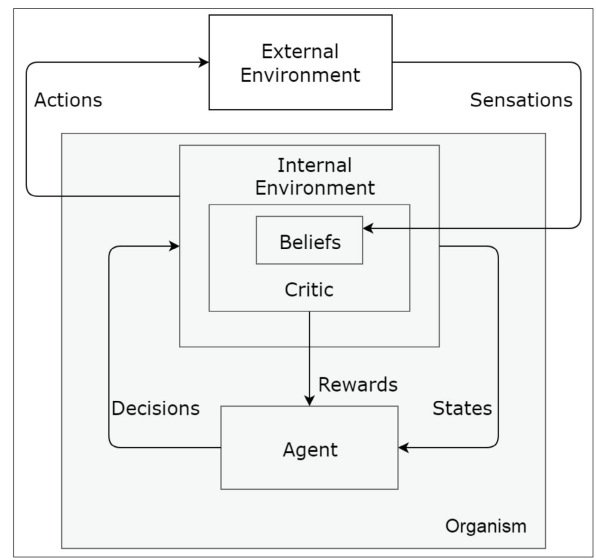

### N. BI-LEVEL OPTIMIZATION OF PARAMETERIZED REWARD SHAPING (BiPaRS)

BiPaRS [10] is a novel approach to address the challenge of effectively utilizing shaping rewards in reinforcement learning. Recognizing that human-designed reward functions can be imperfect due to cognitive biases, BiPaRS employs a bi-level optimization framework. This framework learns to adaptively utilize shaping rewards by optimizing a parameterized weight function for the shaping reward at the upper level, while simultaneously optimizing the policy using the shaped reward at the lower level. This reward shaping technique modifies the original reward function by adding a parameterized shaping term. This technique is represented as follows

$$\tilde{r}(s, a) = r(s, a) + z\phi(s, a)f(s, a), \quad (20)$$

where $\tilde{r}(s, a)$ denotes the shaped reward for taking action $a$ in state $s$. $r(s, a)$ is the original reward function.

$z\phi(s, a)$ is the shaping weight function, parameterized by $\phi$, which assigns a weight to each state-action pair and is parameterized by $\phi$, and $f(s, a)$ represents the shaping reward function and performs a bi-level optimization for $\tilde{r}(s, a)$, which is a nested optimization problem. The aim is to improve an agent's behavior by modifying the reward function it receives. We represent the agent's policy as $\pi_\theta$, where $\theta$ denotes the policy's parameters. The learning objective is twofold, first optimizing the Policy The policy $\pi_\theta$ is optimized with respect to a modified reward function $\tilde{r}$. The goal is to maximize the expected cumulative modified reward:

$$\tilde{J}(\pi_\theta) = E_{s \sim \rho_\pi, a \sim \pi_\theta}[r(s, a) + z_\phi(s, a)f(s, a)]. \quad (21)$$

Second, optimizing the Shaping Function: The shaping function $z_\phi$ is optimized to ensure that the policy $\pi_\theta$ which maximizes $\tilde{J}(\pi_\theta)$ also maximizes the expected cumulative true reward $J(z_\phi)$:

$$J(z_\phi) = E_{s \sim \rho_\pi, a \sim \pi_\theta}[r(s, a)]. \quad (22)$$

This ensures that the shaping reward helps guide the agent towards maximizing the true reward, even though $z_\phi$ itself doesn't directly control the agent's actions.

Therefore, we have a bi-level optimization, an embedded (nested) optimization problem, to optimize the policy $\pi_\theta$ and the weight function $z_\phi$, as follows

$$\max_{\phi} \; \mathbb{E}_{s \sim \rho^\pi, a \sim \pi^\theta}[r(s, a)]$$
$$\text{s.t.} \;\; \phi \in \Phi$$
$$\theta = \arg\max_{\theta^0} \mathbb{E}_{s \sim \rho^\pi, a \sim \pi^{\theta^0}}[\tilde{r}(s, a)]$$
$$\text{s.t.} \; \theta \in \Theta, \quad (23)$$

where $\Phi$ and $\Theta$ represent the parameter spaces of the shaping weight function $z_\phi$ and the policy $\pi_\theta$, respectively.

This method allows BiPaRS to identify and exploit beneficial shaping rewards while mitigating the impact of unhelpful or even detrimental ones. The paper demonstrates BiPaRS's effectiveness through experiments on cartpole and MuJoCo tasks, Figure 20, showing its ability to leverage beneficial shaping and suppress negative influences. However, its performance on MuJoCo tasks may not yet match the cutting edge in the DRL domain. BiPaRS holds promise for adapting and shaping rewards dynamically, but further research is needed to enhance its performance and explore its broader applicability.

## O. REWARD SHAPING VIA HUMAN FEEDBACK

Human feedback serves as a powerful mechanism for enhancing reinforcement learning (RL) by effectively shaping reward structures and accelerating the learning process. A burgeoning area of research within RL involves the integration of human preferences to leverage this feedback, guiding agents toward desired behaviors and objectives. This method seeks to bridge the gap between the agent's learning processes and human intuition, thereby fostering a more effective and efficient learning environment. Notable techniques, such as PEBBLE [67], SURF [68], RUNE [40], and others [69], [70], [71], exemplify the application of Human-in-the-Loop (HITL) methodologies [72], [73], [74], [75], [76] to learn reward functions based on human input.

Reward shaping with human feedback is a specific application of the HITL concept, where human input is used to modify or design reward functions that the RL agent uses to evaluate its actions. This method aims to influence the agent's learning trajectory by providing more informative and context-specific rewards that align with human preferences. By utilizing human feedback, the agent can learn more effectively, as the rewards better reflect the complexities of the tasks at hand.

While both HITL and reward shaping with human feedback utilize human input to improve RL, they differ in focus and application:

- **Scope of Human Interaction:** HITL encompasses a broader interaction framework where human operators provide ongoing feedback, corrections, and guidance throughout the learning process. Reward shaping specifically targets the design and modification of reward functions based on human input, focusing primarily on the reward structure rather than the overall learning process.

- **Learning Dynamics:** HITL fosters a dynamic relationship between humans and agents, facilitating real-time adjustments that can refine learning strategies on the fly. Reward shaping is typically more static, where feedback is incorporated into the reward function design, although it can also involve iterative refinements based on human input.

- **Objectives:** The primary objective of HITL is to enhance the learning efficiency and effectiveness of agents by leveraging human intuition and expertise. Reward shaping aims to provide agents with clearer and more relevant rewards, thereby accelerating the convergence of learning algorithms toward optimal behaviors.

One could model human preferences as a probabilistic distribution over reward functions. For example, the likelihood of a reward function $R$ given human feedback $\mathcal{H}$ can be modeled as:

$$P(R|\mathcal{H}) = \frac{P(\mathcal{H}|R)P(R)}{P(\mathcal{H})}. \quad (24)$$

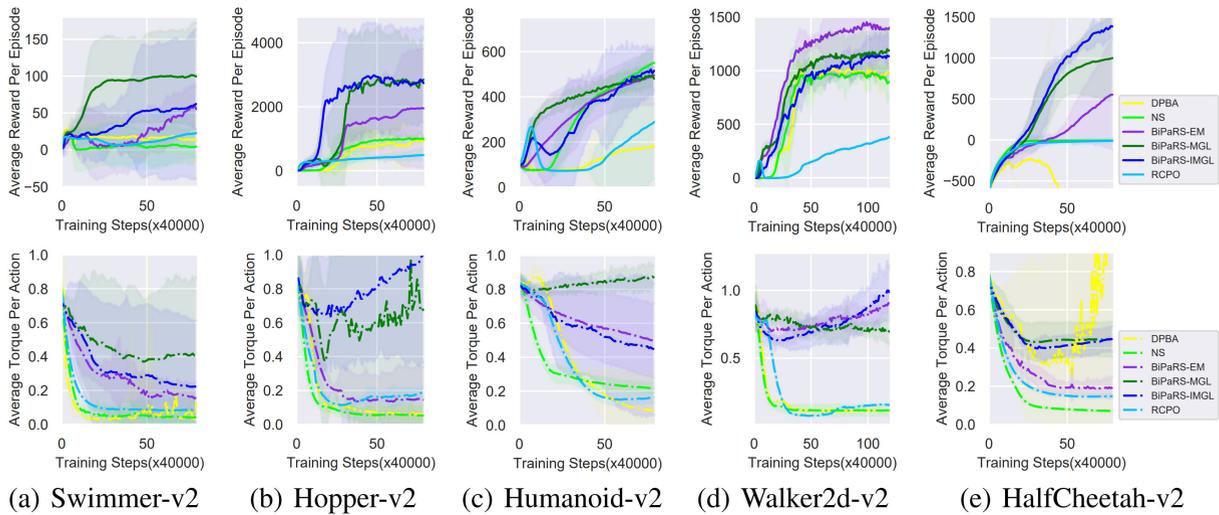

**FIGURE 20.** Results of the MuJoCo experiment, where the shaded areas are 95% confidence intervals [10, Figure 19].

Here, $P(\mathcal{H}|R)$ represents how likely the human feedback is given a specific reward function $R$, and $P(R)$ represents a prior distribution over reward functions.

By maximizing the posterior probability $P(R|\mathcal{H})$, one can derive a reward function that aligns well with human preferences, thereby improving the learning efficiency and effectiveness of the RL agent. One of the methods that incorporates large-scale language models (LLMs) is Text2Reward, it uses LLMs to automatically create dense reward functions for reinforcement learning tasks, thereby reducing the reliance on domain expertise or extensive data collection [77]. This approach has been practically applied to a robotic arm performing two specific tasks: Pick Cube (i.e., picking up a cube and moving it to a designated position) and Stack Cube (i.e., stacking one cube onto another), as illustrated in Figure 21.

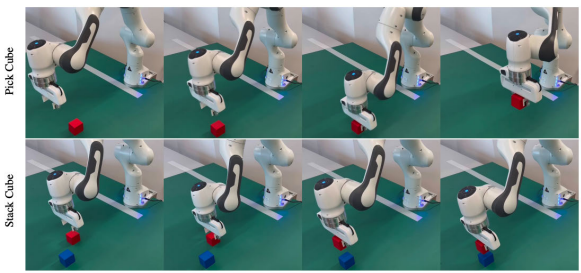

**FIGURE 21.** Sample images of real-world robot manipulation for the Pick Cube and Stack Cube tasks, [77, Figure 5].

By utilizing natural language descriptions of desired objectives, Text2Reward generates reward functions that are both interpretable and executable. These automatically generated rewards have shown superior performance in various robotic manipulation tasks compared to manually designed ones. The method's effectiveness stems from its ability to produce adaptable dense rewards, which have been proven successful in both simulated and real-world robotic manipulation and locomotion tasks. However, Text2Reward currently faces challenges in handling highly complex tasks. Although the primary focus is on robotics, the authors anticipate broader applications in areas such as gaming, web navigation, and household management, where automated reward design could significantly accelerate the creation of intelligent systems.

This work [78] proposes a Meta-Reward-Net (MRN), a new framework for preference-based reinforcement learning that leverages Preference-based reinforcement learning through human feedback to improve learning efficiency. The gap between MRN and baselines diminishes as the amount of feedback increases, suggesting that sufficient feedback alleviates the limitations imposed by insufficient preference labels. Furthermore, Bi-level optimization, as we've mentioned can be an efficient tool for Reward Engineering, and when combined with reward learning and limited human feedback, it can generate an accurate Q-table, compared with the State-of-The-Art methods.

MRN employs bi-level optimization to simultaneously learn both reward functions and policies. This dual approach allows the agent to learn effectively, even with constrained human input, outperforming previous methods across a variety of simulated robotic tasks. The authors assert that traditional reward shaping methods are inadequate for achieving robust and efficient reinforcement learning. In summary, MRN enhances feedback efficiency by incorporating an additional signal that considers the performance of the Q-function based on labeled preference data.

Experiments conducted on Meta-world tasks illustrate MRN's superior performance with limited feedback.

However, the reliance on human-guided reward engineering, which prioritizes exploration, stability, and context-specific design, presents a potential drawback, as its efficacy depends heavily on the quality of human input. The paper emphasizes the critical importance of iterative refinement and human feedback in overcoming challenges such as sparse, delayed, and noisy rewards. While the primary focus of this research is on robotic simulations, it is noteworthy that the code for MRN is publicly available, enabling further implementation and testing by the research community, thereby facilitating a continued exploration of HITL methodologies in RL contexts.

### P. UNDERREPRESENTED REWARD SHAPING TECHNIQUES AND METHODS

Every RL method has a distinct reward shaping technique, parameterized values, design methods, algorithm, or environment, and all of them share the existence of a reward function, to cover these issues we need to take a deeper look into the literature, and that could not be feasible with the resources at hand. Nonetheless, we will name a few of the underrepresented methods that were used to shape or design the reward:

#### 1) REWARD SHAPING FOR DEEP Q-LEARNING IN POWER GRID CYBERATTACK DEFENSE

Deep Q-learning is a powerful method in RL but one of its issues is that it might not be compatible with large environments with high dimensions or a large state space. Nevertheless, it can be used to address critical issues such as the protection and security of smart power grids against cyberattacks [79], [80]. To overcome these challenges, the authors of [79] introduce a deep Q-learning-based stochastic zero-sum Nash strategy solution. This approach utilizes a deep neural network to learn a control policy that minimizes the impact of attacks while considering the probabilistic nature of attack outcomes, and it uses reward shaping to find scenarios where the attacker can harm the network the most, and therefore find the best defense strategies.

The reward function [79, Equation 10] is designed to incentivize the attacker to cause as much damage as possible by targeting transmission lines. It assigns rewards based on the following criteria:

- Exceeding the Attack Objective (AO): If the attack causes more immediate outages (IO) than the defined attack objective (AO), the attacker receives a high reward ($r_1$).
- Reaching the Attack Objective: If the attack successfully reaches the attack objective, the attacker receives a lower reward ($r_2$).
- Partial Success: If the attack causes some outages but doesn't reach the objective, the reward is proportional to the ratio of immediate outages (IO) to the attack objective (AO).

The reward values ($r_1$ and $r_2$) are chosen such that $r_1$ is significantly larger than $r_2$, encouraging the agent to learn actions that maximize immediate damage and exceed the attack objective whenever possible.

The results demonstrate that the proposed deep Q-learning algorithm significantly outperforms the alternative RL algorithms, achieving a higher level of defense against attacks. While the deep Q-learning algorithm may require more time to converge, its superior reliability and efficiency compensate for the slower convergence, particularly in scenarios where real-time computation is not critical.

### Q. OTHER METHODS

Various innovative reward-shaping frameworks enhance RL approaches across diverse challenges and environments. One approach employs barrier functions (BFs) to ensure safe RL agent behavior, demonstrating accelerated convergence and reduced actuation effort in simulations and real-world deployments [81]. Another method utilizes natural language instructions to generate dense rewards, improving RL efficiency while posing challenges in natural language processing integration [82]. Additionally, temporal logic-based reward shaping for average-reward RL leverages formal logic to enhance learning rates [83]. Addressing RL over-optimization, a framework penalizes rewards based on uncertainties, while another focuses on using reward expectations to stabilize and accelerate convergence in uncertain RL environments [13], [84].

In a reward planning scheme that relies on the tessellation of the state space, and dividing the tasks into sub-tasks, then planning the agent's behavior through the state space, the works [85], [86] propose a novel reward planning method, "Greedy Divide and Conquer". This method is designed for underactuated robotic systems operating under parameter uncertainty. The method utilizes a single reinforcement learning (RL) agent to address the challenge of swing-up and balancing a Pendubot system with uncertain parameters. This approach allows the agent to adapt to parameter uncertainty and achieve faster convergence compared to using a non-shaped reward function. The method was tested on a Pendubot system with uncertainties up to 200-300% in various parameters, achieving an average 95% accuracy across trials. The method's strengths lie in its ability to handle significant uncertainties and design reward functions systematically by breaking down complex tasks into sub-problems. It can also be used to avoid specific actions or behaviors during task execution. However, the method's downside is the potential for time-consuming development of the reward function, requiring a thorough understanding of the system.

#### 1) AUTOMATED PARAMETER TUNING

While deep reinforcement learning has made substantial progress determining optimal hyperparameters and reward functions remains a significant challenge even for experienced practitioners many studies rely on established benchmarks where prior knowledge about these crucial aspects is readily available however real-world applications frequently involve novel and intricate tasks lacking such pre-existing knowledge necessitating the development of these components from scratch.

This research [87] proposes a novel approach to address this challenge by jointly optimizing and combining hyperparameter auto-tuning and reward functions. The authors recognize that these components are intrinsically linked and cannot be effectively optimized in isolation to facilitate this optimization they utilize DEHB [88], a cutting-edge hyperparameter optimization algorithm DEHB combines the strengths of HyperBand optimization with Differential Evolution, functioning like genetic algorithms DEHB has consistently demonstrated robust performance across a variety of benchmarks.

The authors conducted experiments using proximal policy optimization (PPO) and soft actor-critic (SAC) algorithms in diverse environments the results demonstrate, as can be seen in Figure 22, that this joint optimization strategy significantly improves performance compared to optimizing each component individually this research highlights the potential benefits of this method suggesting it may become a standard practice for RL optimization. However, the research acknowledges limitations, such as the focus on optimizing specific reward parameters within a predefined structure, suggesting further exploration of broader reward function combinations, and it mentions the need for further research to explore a broader range of reward function configurations and investigate more sophisticated risk-averse metrics to achieve a balance between performance and stability.

To delve deeper into the topic, a recent review provides a comprehensive review of the field of automated reinforcement learning (AutoRL) [89], aiming to automate key components of the RL framework making it accessible even to non-experts. The paper proposes a general AutoRL pipeline breaking down the framework into three crucial components MDP modeling, algorithm selection, and hyperparameter optimization.

The review highlights the significant advantages of AutoRL, including reducing the need for specialized RL expertise automating time-consuming tasks like defining state and action spaces, and potentially leading to more efficient and effective RL solutions. However, the paper acknowledges that AutoRL is still a relatively nascent research area and faces several challenges. One key limitation is the lack of a concrete and comprehensive AutoRL, pipeline akin to the well-established automl pipelines. Moreover, optimizing hyperparameters efficiently automatically modeling problems as MDPS and generalizing the mapping between information and RL environments remain critical research questions.

The importance of hyperparameter optimization is underscored in the paper recognizing that achieving optimal RL configurations for solving sequential decision-making problems hinges on carefully chosen hyperparameter settings, these hyperparameters fixed during training are typically set by RL experts before the training phase, examples include learning rates in policy gradient, or value function updates, discount factors, eligibility trace coefficients, and parameters for parametric reward shaping methods.

Different tasks often require distinct sets of hyperparameters, making hyperparameter optimization a challenging endeavor. Automating this process, as the paper suggests, would be extremely beneficial. The research explores various methods for hyperparameter optimization, including stochastic gradient descent and neural networks, bayesian optimization, multi-armed bandit, evolutionary algorithms, greedy algorithms,s and finally reinforcement learning itself.

Many approaches developed for automatically optimizing hyperparameters in supervised learning algorithms can be adapted to optimize hyperparameters of RL algorithms. This section further reviews hyperparameter optimization techniques and their application in RL. Lastly, the main challenges of optimizing hyperparameters within the context of AutoRL are discussed. The paper emphasizes the potential benefits of AutoRL for solving complex sequential decision-making problems across various domains, potentially leading to significant time and resource savings. The integration of automated hyperparameter optimization into the AutoRL pipeline promises to further enhance its capabilities and contribute to more efficient and robust RL solutions.

## V. REAL WORLD APPLICATIONS
This section delves into the specific challenges and opportunities presented by reward engineering in robotics. We will explore notable robotic applications across various domains, highlighting how these applications relate to the reward engineering techniques discussed in Section IV. This will aim to provide a practical context for understanding the implications or reward shaping within real-world robotic applications.

### A. REWARD SHAPING FOR SAFE AND EFFICIENT HUMAN-ROBOT COLLABORATION
To enhance safe interactions between humans and robots in industrial settings [90] presented a deep reinforcement learning (DRL) approach, the study focused on enabling robots to autonomously learn policies that minimize risks and optimize task efficiency. The framework of Human Robot Collaboration (HRC) is illustrated in Figure 23. Central to the approach is the development of the Intrinsic Reward-Deep Deterministic Policy Gradient (IRDDPG) algorithm, which integrates deterministic policy gradient (DPG) methods with an optimized reward function combining intrinsic and extrinsic rewards.

Experimental results demonstrate that IRDDPG enables robots to learn collision-avoidance policies effectively, ensuring safety in human-robot interactions while achieving task objectives. This method shows better results than manually

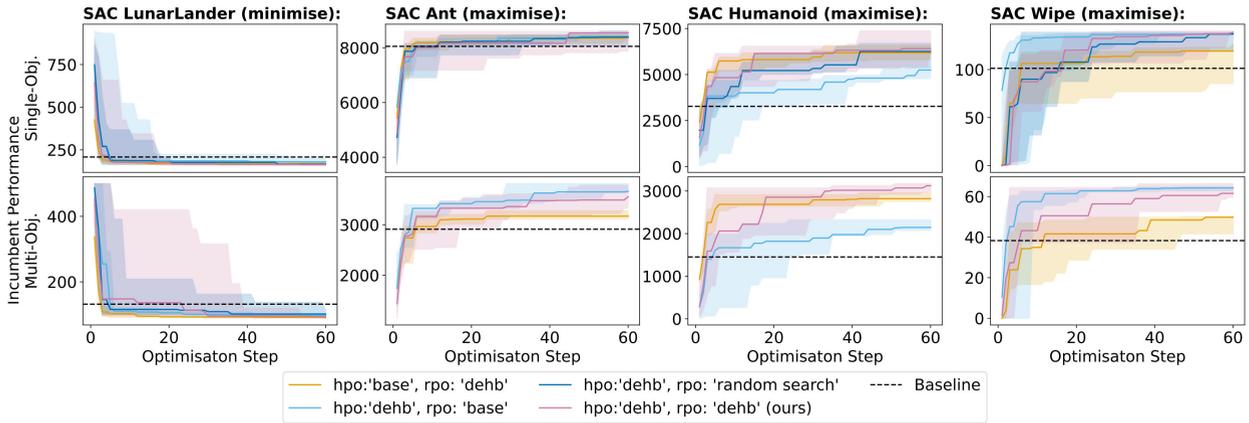

**FIGURE 22.** Median optimization objective for SAC (5 optimization Runs), shaded areas indicate Min/Max values [87, Figure 3].

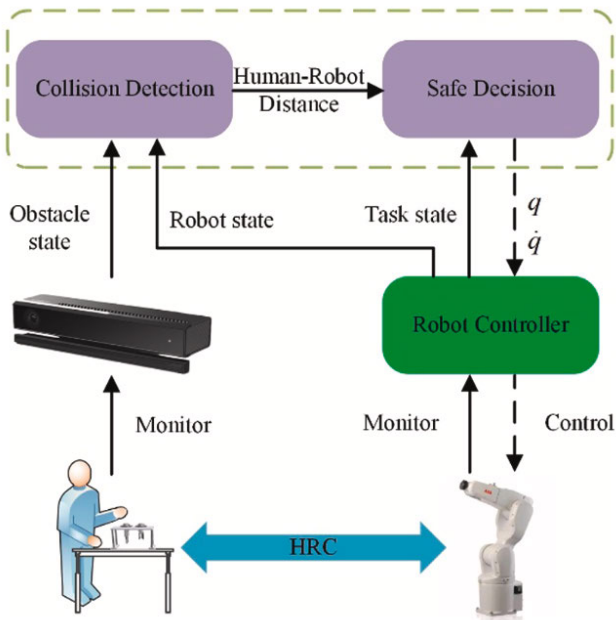

**FIGURE 23.** The framework of collision avoidance in safe HRC, [90, Figure 2].

The schematic is illustrated in Figure 24, comparing their effectiveness in accelerating learning and reducing errors.

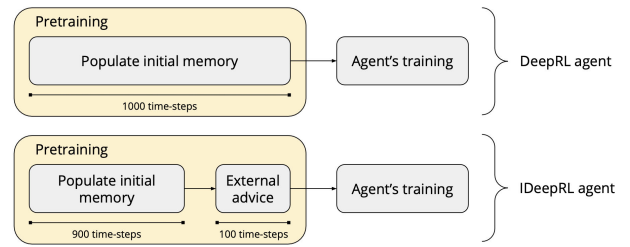

**FIGURE 24.** The learning process for autonomous and interactive agents. Both approaches include a pretraining stage comprising 1000 actions. For interactive agents, the final part of the pretraining is performed using external advice instead of random actions, [91, Figure 2].

Results demonstrate that interactive feedback from both human trainers and artificial agents significantly improves performance metrics such as total collected rewards and task completion speed compared to autonomous DeepRL. Human feedback exhibits slightly superior results in certain scenarios, highlighting the value of incorporating human expertise into robotic learning processes. This approach emphasizes the optimization of feedback strategies to guide agent behavior toward achieving specific task objectives efficiently, which enhances learning efficiency and effectiveness through interactive feedback mechanisms. However, challenges may involve optimizing these feedback strategies for diverse real-world applications.

A distributed multi-robot navigation strategy, which merges Reciprocal Velocity Obstacles (RVO) with DRL, addresses the challenges of collision avoidance in complex environments with limited information [93]. This method, referred to as RL-RVO, seamlessly integrates RVO concepts with learning techniques. It utilizes sequential velocity obstacle (VO) and RVO vectors to model the environment, employing a bidirectional recurrent neural designed reward functions by dynamically adjusting rewards during learning, it also demonstrated the efficacy of tailored reward structures in guiding agent behavior towards desired safety and efficiency outcomes, which improved safety through autonomous collision avoidance and task completion efficiency, yet challenges may arise in the complexity of reward function optimization and computational demands.

In the same context, [91] enhances DRL for domestic robots performing household tasks like organizing objects with a robotic arm by exploring the application of interactive feedback. The study evaluates three learning methods: autonomous DRL, agent-assisted interactive DRL (agent–IDRL), and human-assisted interactive DRL (human–IDRL), network to translate obstacle states into robot actions. The reward function is meticulously designed to balance collision avoidance with travel time efficiency, thereby optimizing the navigation performance of the robots. Simulation evaluations with varying numbers of robots and obstacles show that this approach outperforms other state-of-the-art methods in terms of success rate, travel time, and average speed. The RL-RVO policy is further implemented and tested on Turtlebot robots to validate its real-world performance, as illustrated in Figure 26.

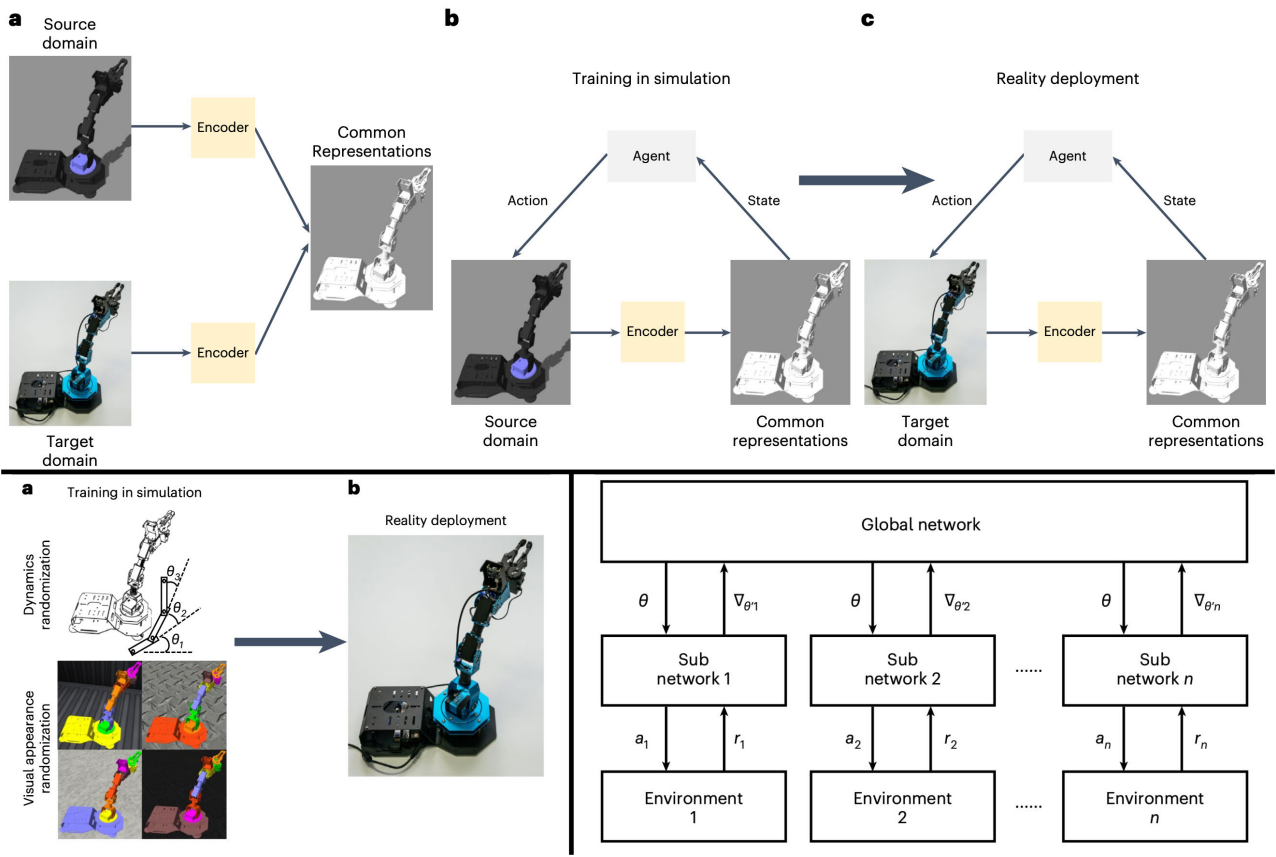

**FIGURE 25.** Top: *Domain Adaption* [92, Figure 1]: a) An encoder learns to map both the simulated (source) and real (target) environments to a shared state space. b) This shared state space is then used to train the policy within the simulation. c) The policy trained in the simulation is transferred to the real environment, with states being mapped to the shared representations for further learning.
Bottom Left: *Domain Randomization* [92, Figure 2]: a) The agent's training experience is enhanced by randomizing the physical dynamics and/or visual appearance within simulated environments. b) The simulation-trained policy is expected to perform well in real-world tasks after a one-shot transfer.
Bottom Right: *Meta RL Learning* [92, Figure 4]: The overall policy network is transferred to a set of subpolicy networks, which interact with a batch of sampled meta-training tasks (environments 1, 2, ..., n). The subpolicy network is updated based on rewards from the corresponding task, and the global network is optimized using the parameters of the subpolicy networks.

These experiments involve up to eight differential drive Turtlebots, all arranged in a circular formation with random orientations. In this circle scenario, each robot, starting with a random orientation, is uniformly placed around the circle's perimeter, with the goal position located on the opposite side. This setup creates a rich interaction environment for the robots, as shown in Figure 27.

By integrating Multifunctional Reward Shaping, which guides the robot towards its destination while simultaneously avoiding obstacles and providing informative rewards that accelerate learning, with Hindsight Experience Replay (HER), which mitigates the challenge of sparse rewards by allowing the robot to learn from both successful and unsuccessful experiences, [94] proposes an end-to-end RL strategy for navigating autonomous mobile robots in dynamic environments. This combined approach enables the robot to develop an optimal navigation policy, showing a strong capacity to adapt to previously unseen scenarios.

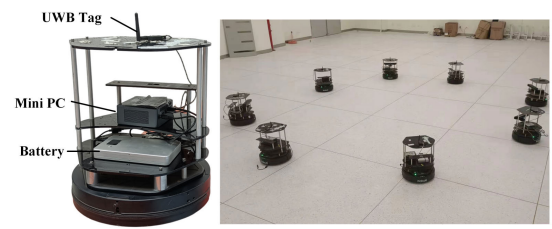

**FIGURE 26.** Illustration of real-world experiments: (left) a single Turtlebot, (right) eight Turtlebots uniformly positioned in a circle with random orientations [93, Figure 7].

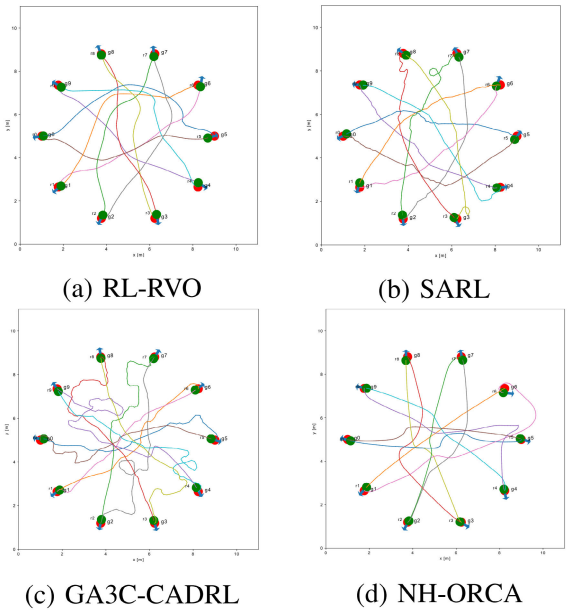

**FIGURE 27.** Illustration of trajectories generated by four different approaches in the circle scenario [93, Figure 3].

The effectiveness of this method has been demonstrated through both simulations and real-world experiments. In the training environment, the destination is randomly assigned at the start of each run, with the robot's starting point at the center of the space. Upon reaching the destination, the navigation resets from that point. Dynamic obstacles are represented by four cylindrical structures that rotate with a fixed radius. The Gazebo simulation platform allows for the creation of environments that closely resemble real-world conditions, thereby reducing development time and costs while enhancing convenience, as shown in Figure 28.

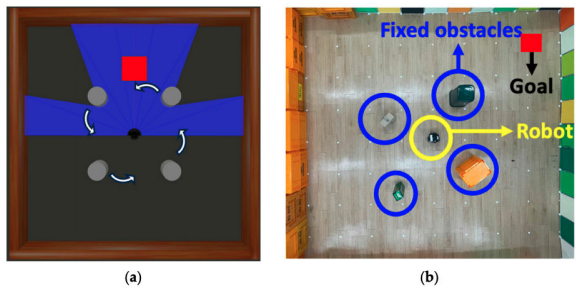

**FIGURE 28.** (a) The test-driving environment in simulation, (b) the real-world test-driving environment, distinct from the trained environment [94, Figure 10].

### B. REWARD SHAPING FOR AUTONOMOUS VEHICLES AND TRAFFIC FLOW OPTIMIZATION

One promising direction in the development of autonomous vehicles is the application of off-policy reinforcement learning methods for traffic navigation [95]. A key approach in this area is the Episodic-Guided Prioritized Experience Replay (EPER) method, which addresses the inherent sample inefficiency faced by Deep Reinforcement Learning (DRL) models in autonomous driving applications. Typically, DRL-based navigation requires extensive interactions with the environment to derive optimal policies, making the learning process time-consuming and resource-intensive. EPER alleviates this challenge by utilizing episodic memory to store successful experiences and extract expected returns for each state-action pair, combined with Temporal Difference (TD) error-based prioritization. This dual approach accelerates the training process while maintaining robust policy development.

In addition, EPER integrates a regularization term that fosters the exploration of diverse state-space regions, thereby preventing the learning process from becoming excessively deterministic. When applied to complex traffic scenarios such as highway driving, merging, roundabouts, and intersections, EPER has demonstrated superior performance compared to conventional Prioritized Experience Replay (PER) and other state-of-the-art techniques. Results indicate that EPER enhances sample efficiency, enabling quicker convergence to optimal policies and significantly lowering collision rates in simulated driving environments.

Moreover, by integrating EPER with reward shaping techniques, autonomous vehicles can be trained to optimize their navigation strategies more effectively. Reward shaping can guide the learning agent towards desired behaviors—such as collision avoidance and adherence to traffic rules—by adjusting the reward function. This process can ensure that the agent's learning remains focused on long-term objectives rather than immediate rewards, leading to safer and more reliable driving behaviors.

Beyond autonomous vehicles, there is potential to apply similar reinforcement learning techniques, such as reward shaping, to the optimization of traffic flow in worst-case scenarios. In this context, reward shaping could be utilized to optimize traffic flow independently of time constraints, similar to the way quantum annealing has been employed to solve complex optimization problems related to traffic flow and congestion, as demonstrated in previous works [96], [97]. These studies highlight the potential of quantum annealing in optimizing traffic networks, but the authors believe that the same problems could be addressed with reward shaping in a reinforcement learning framework. By designing reward functions that prioritize minimizing congestion and enhancing the overall efficiency of traffic systems, reinforcement learning could offer a practical and scalable solution to real-time traffic flow challenges.

Thus, reward shaping, combined with reinforcement learning, presents a promising avenue for not only improving autonomous vehicle navigation but also addressing broader traffic management issues, including optimizing flow under worst-case conditions.

## VI. SIM-TO-REAL IN REWARD ENGINEERING/SHAPING

Reward shaping and engineering are essential for effectively transferring DRL and RL policies from simulations to real-world scenarios. It facilitates this transition by encouraging behaviors that are robust in both simulated and real environments. The connections between various methods for transferring from simulation to reality are illustrated in Figure 29.

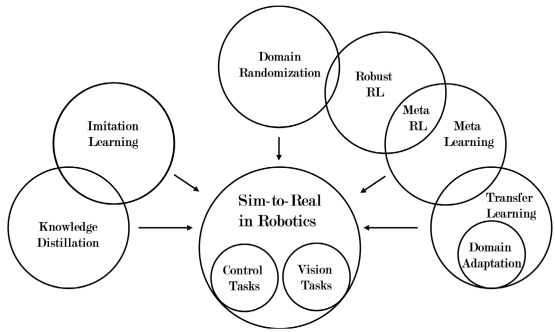

**FIGURE 29. Various techniques for transferring from simulation to reality in deep reinforcement learning and their interconnections [98, Figure 2].**

Figure 30 provides a graphical breakdown of the various components within a RL algorithm. It also highlights the distinct challenges faced when training DRL algorithms. Key obstacles include achieving consistent outcomes across different implementations and datasets, which underscores the issue of reproducibility. The sensitivity of DRL algorithms to hyperparameters adds another layer of complexity, complicating performance optimization. Additionally, the implementation and deployment of DRL algorithms are often intricate and time-consuming tasks [99].

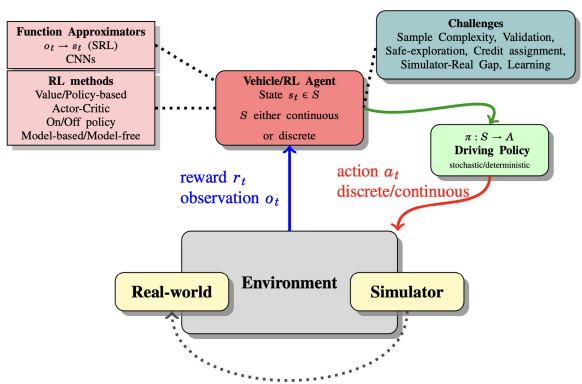

**FIGURE 30. Simulator-to-real gap [99, Figure 3].**

The shift of DRL policies from simulation to real-world robotic applications is challenging due to discrepancies between the two environments, compounded by concerns about safety, cost, and efficiency. Techniques to address this performance gap include domain adaptation, domain randomization, and progressive neural networks (PNNs) [92], [98], [100]. Domain adaptation aims to minimize the differences between simulated and real environments, as illustrated in the upper Figure 25. On the other hand domain randomization enhances policy robustness through varied simulation parameters, illustrated in the lower left Figure 25, PNNs and meta-reinforcement learning further improve transfer efficiency by leveraging knowledge from previous tasks, illustrated in the lower right Figure 25.

To address these challenges, [101] introduced the Consensus-based Sim-And-Real DRL algorithm (CSAR), which exemplifies reward shaping by optimizing policies suitable for both simulation and real contexts, ensuring consistent reward structures throughout different training stages. The CSAR algorithm integrates consensus-based training with DRL in both simulated and real environments, optimizing policies by concurrently training agents in both settings. This consensus-based method, which runs simulated and real agents in parallel, mitigates transition issues and reduces training time. Results indicate that an increased number of agents in simulation benefits both sim-and-real training, as illustrated in Figure 31. Similarly, [102] presents a two-step sim-to-real process that utilizes an intermediate semi-virtual environment. This approach integrates real robot dynamics with simulated sensors and obstacles, as illustrated in Figure 32. This configuration enables the controlled incorporation of real-world complexity while preserving the flexibility of simulation.

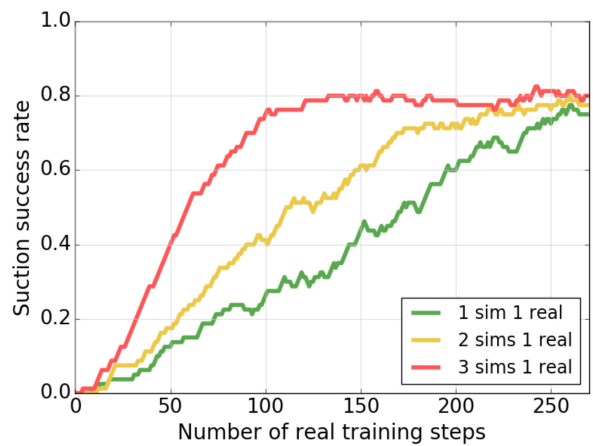

**FIGURE 31. Suction success rates of the real robot with a different number of simulated robots using Sim-and-Real strategy [101, Figure 6].**

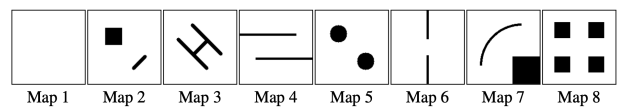

**FIGURE 32. The evaluation maps [102, Figure 3].**

Figure 33 shows the computational graph of the method, featuring an environment interaction thread (left) and a model update thread (right). Both threads interact with the replay buffer, necessitating the use of a mutex to prevent conflicts.

Fortunately, the environment interaction thread adds an entry at the end of its cycle, while the model updates thread samples from it early on, resulting in a low likelihood of mutual blocking. Moreover, parallel data collection and model updates for real-time fine-tuning allow the RL model to adapt to real-world changes without disrupting operations. Deploying the model at a high inference frequency achieves performance nearly equivalent to simulation, even without initial fine-tuning.

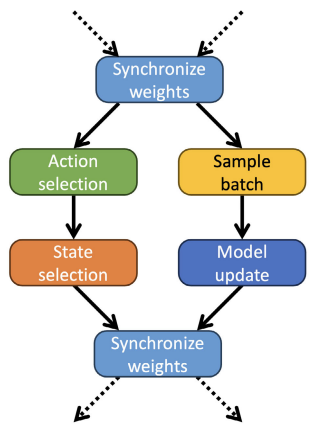

**FIGURE 33.** Environment interaction thread (left) and a model update thread (right) **[102, Figure 1].**

In the same context, rewarding a cleaning robot for maintaining a mess-free environment can lead to reward hacking [103]. In such scenarios, the robot might disable its vision to avoid detecting messes, cover messes with opaque materials, or hide when humans are around to prevent them from pointing out new types of messes. Reward hacking occurs when agents exploit loopholes in the reward function to achieve high rewards without actually fulfilling the intended tasks.

Integrating reward shaping with advanced techniques like domain adaptation, domain randomization, meta-learning, and consensus-based training is essential for overcoming challenges in transferring DRL and RL policies from simulation to real-world robotics. These approaches collectively strengthen policy robustness and efficiency, facilitating their effective deployment in real-world applications. Continuous enhancements in simulation quality, efficient learning strategies, and well-defined reward structures are pivotal for advancing the field and achieving dependable performance of DRL and RL systems in real-world settings. Furthermore, ensuring effective reward shaping is crucial to prevent reward hacking, ensuring that rewards promote desirable behaviors in both simulated and real environments.

## VII. ADVANTAGES AND DISADVANTAGES OF REWARD ENGINEERING: IS IT THE FUTURE?

Reward shaping/engineering is a powerful technique used to enhance the performance of RL agents. It can significantly improve the effectiveness of RL algorithms in several ways. By providing additional information through shaping, agents can learn optimal policies much faster, and this leads to accelerated learning. This is especially beneficial in complex environments where traditional RL methods might struggle to converge quickly. This is highlighted in discussions of STRIPS-based shaping, UCBVI-shaped, DRiP, PBRS, and other methods. Reward shaping can incentivize agents to explore more effectively, particularly in environments with sparse rewards, and therefore improve exploration. This allows agents to discover valuable states and actions that might otherwise go unnoticed. Reward shaping can sometimes lead to agents achieving higher performance levels than they would without shaping, as was discussed in Section IV, and other methods demonstrating performance gains.

Reward shaping can improve the robustness of learned policies, making them less susceptible to noise, uncertainty, or changes in the environment.

Despite its numerous benefits, reward shaping also poses certain challenges, many reward shaping methods rely heavily on domain knowledge to design effective shaping functions. This can be a significant limitation, especially in complex or unfamiliar environments where expert knowledge might be scarce or difficult to acquire.

Furthermore, computational complexity could be considered a disadvantage for some reward shaping methods, especially those involving potential-based shaping, which can increase the computational burden on the learning algorithm. This can be a concern in real-time applications or with limited computational resources, but this drawback is only for some methods, other methods can decrease the computational complexity. The effectiveness of many reward shaping techniques depends on carefully tuned parameters, although some approaches IV-N can tune their parameter, but finding the optimal parameter settings can be a challenging process, and incorrect parameter values can negatively impact performance. Finally, some reward shaping techniques require careful designing and shaping reward functions, and this can be a time-intensive process.

In conclusion, reward shaping is a valuable tool in the reinforcement learning toolbox. While it offers many advantages, careful consideration must be given to its potential drawbacks. Choosing the right reward shaping method is crucial for successful application.

## VIII. OPEN CHALLENGES AND FUTURE DIRECTIONS

Future research directions in robotic manipulation, pivotal across sectors such as manufacturing, healthcare, and space exploration, especially in Industry 4.0 contexts, involve enhancing sample efficiency, bolstering algorithm robustness, promoting human-robot collaboration, and investigating advanced neural network architectures [104]. The importance of reward shaping and engineering in optimizing learning outcomes within Deep Reinforcement Learning (DRL) underscores potential avenues for further advancements in this vital domain.

For tasks involving image processing or complex sensor data where designing rewards is difficult, we find that end-to-end RL approaches can tackle these tasks, such as the research [14] which uses end-to-end RL combining deep neural networks with SAC, enabling robots to learn from a limited number of successful demonstrations. This reduces the reliance on explicit reward functions, making it particularly beneficial for tasks involving image processing or complex sensor data where designing rewards is difficult. The method was evaluated on robotic manipulation tasks, such as picking and placing objects, and the authors claim to achieve high success rates and accuracy (100%).

However, the research has several limitations. Firstly, the robot's behavior may not always be optimal, potentially leading to sub-optimal solutions or inefficient movements. Secondly, the method is not robust to uncertainties in the environment, such as variations in object placement or lighting conditions. Thirdly, the reliance on user queries can potentially be time-consuming, especially for tasks with a large number of possible actions or complex decision-making processes.

## IX. CONCLUSION

This work provides a comprehensive review of reward shaping techniques within the field of RL. A detailed taxonomy of methods is presented encompassing descriptions, advantages, application domains, and relevant metrics. This analysis addresses open challenges and future directions, offering valuable insights for future research and development. Our findings highlight the significant benefits of reward shaping, including its ability to highly expedite learning, manage uncertainties, bolster robustness, enhance system outcomes, and increase the success rate of RL agents. While reward shaping demonstrably improves learning outcomes, its implementation can be complex and time-consuming, particularly in scenarios with significant domain complexity. Automated parameter tuning methods mitigate this challenge to some extent. Further research is required to evaluate reward shaping techniques in real-world applications and explore the potential of incorporating human feedback into the training process. This work aims to establish a comprehensive resource for understanding reward shaping in RL. This resource facilitates the selection of appropriate methods based on the specific problem, enabling researchers and practitioners to effectively leverage the power of reward shaping for improved RL performance.

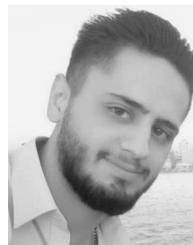

**SINAN IBRAHIM** received the B.S. degree in mechatronics engineering from Tishreen University, Syria, in 2021, and the M.S. degree in robotics and computer vision from Innopolis University, Russia, in 2023. He is currently pursuing the Ph.D. degree in computational and data science and engineering with Skolkovo Institute for Science and Technology. He is a highly accomplished Researcher and an Engineer with expertise in robotics, computer vision, and reinforcement learning. He has a proven track record of developing innovative solutions and is passionate about both scientific advancement and its real-world business applications. His career has involved a blend of teaching, coaching, business, fundraising, go-to-market, and hyper-growth for startups, demonstrating his adaptability and diverse skill set. His research interests include computer vision, robotics, and data-based control.

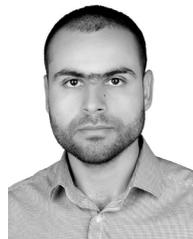

**MOSTAFA MOSTAFA** received the bachelor's degree in mechanical and electrical engineering from the Mechatronics Engineering Department, Tishreen University, Latakia, Syria, in 2017, and the master's degree (Hons.) in intelligent technology in robotics from ITMO University, Saint Petersburg, Russia, in 2020. He is currently pursuing the Ph.D. degree in reinforcement learning and robotics with Skolkovo Institute of Science and Technology, Moscow, Russia. He received certificates in various fields, advances in robotics from LUT University, Finland, and ITMO University, Russia; the elements of AI (artificial intelligence) from the University of Helsinki, Finland; and professional development (active learning in STEM) from Skolkovo Institute of Science and Technology. From 2020 to 2022, he was a Quality Engineer (a Process Engineer) in Saint Petersburg, where he established methods for solving technical problems, conducted performance loss studies, systematized maintenance tools, and implemented production management tools. Additionally, he identified key projects to achieve performance targets and collaborated with technical experts and other manufacturing sites to create optimal solutions.

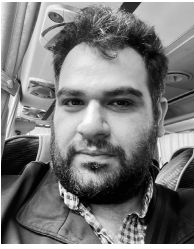

**ALI JNADI** received the B.Sc. degree in mechatronics from Tishreen University, Syria, in 2011, and the M.Sc. degree in robotics and computer vision from Innopolis University, Russia, in 2023, where he is currently pursuing the Ph.D. degree, focusing on advanced control systems for underactuated systems.

From 2013 to 2018, he taught informatics at the National Center for Distinguished, Syria. From 2018 to 2019, he was a Specialist Instructor with the Ministry of Education, Syria. From 2019 to 2021, he taught and supervised the Robotics and Control Laboratory, Manara University, Syria. Since August 2022, he has been a Teaching Assistant with Innopolis University, teaching labs in differential equations, data structures and algorithms, human-AI interaction design, and physics. Since November 2023, he has been an Engineer with Innopolis University's Technology Center for Robotics and Mechatronics Components, working on local planning and control algorithms for autonomous vehicles. Since 2023, he has been a Researcher on a project called Robotic Incremental Metal Sheet Forming. His research interests include model predictive control, reinforcement learning, and robotics, with published works on explicit model predictive control design and reinforcement learning methods for "Pendubot" balancing.

Mr. Jnadi has received several awards, including the First and Second Places in High School Sumo Competitions at ARC7 Syria, in 2022, the Second Place in the Future Engineer Category at the WRO 2021 Finals Future Engineer Category, and the Gold Medal at The "Al Bassel" Fair for Invention and Innovation for his Hybrid Motorcycle Project, in 2011.

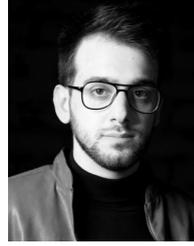

**HADI SALLOUM** is currently pursuing the bachelor's degree in artificial intelligence and data science, with a focus on applying advanced AI techniques to complex problems across various domains. He has demonstrated exceptional commitment to physics, competing in the International Physics Olympiad 2021 in Lithuania and European Physics Olympiad in Estonia. He is a Researcher with the AI Institute, Innopolis University, Russia. His research interests include innovative AI and data science applications, particularly utilizing quantum computing to enhance computational capabilities.

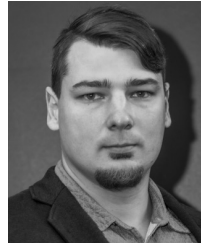

**PAVEL OSINENKO** received a diploma with honors from Bauman Moscow State Technical University in 2009, and the Ph.D. degree from Dresden University of Technology, in 2014. From 2011 to 2020, he worked at German academic and industrial sectors focusing on research, project supervision, and teaching. Since 2020, he has been an Assistant Professor with Skoltech, Moscow. His research interests include reinforcement learning, nonlinear system stabilization, and computational aspects of dynamical systems.

• • •